\documentclass[5p,times]{elsarticle}

\usepackage{hyperref}

\usepackage{amsmath}
\usepackage{amssymb}
\usepackage{multirow}
\usepackage{diagbox}
\usepackage{colortbl}

\usepackage[english]{babel}
\usepackage{blindtext}
\usepackage[utf8]{inputenc}
\usepackage[T1]{fontenc}
\usepackage{hyperref}
\usepackage{subcaption}
\usepackage{float}

\usepackage{xcolor}
\usepackage{multirow}
\usepackage{caption}
\usepackage{indentfirst}

\DeclareMathOperator*{\concat}{||}

\usepackage{xcolor}

\newcommand{\R}{\mathbb{R}}

\newcommand*{\figuretitle}[1]{%
    {\centering
        \begin{footnotesize}
            \vspace{-.1em}#1\par\medskip\vspace{-.4em}
        \end{footnotesize}
    }
}

\makeatletter
\newcommand\footnoteref[1]{\protected@xdef\@thefnmark{\ref{#1}}\@footnotemark}
\makeatother

\journal{arXiv}

\begin{document}\sloppy

\begin{frontmatter}

\title{TENT: Tensorized Encoder Transformer for temperature forecasting}

\author[]{Onur Bilgin}
\author[]{Paweł Mąka}
\author[]{Thomas Vergutz}
\author[]{Siamak Mehrkanoon\corref{corauthor}}

\address{Department of Data Science and Knowledge Engineering, Maastricht University, The Netherlands.}
\cortext[corauthor]{Corresponding author. S. Mehrkanoon is also with Mathematics Centre Maastricht, Maastricht University, The Netherlands.\\
(e-mail: siamak.mehrkanoon@maastrichtuniversity.nl).}

\begin{abstract}
\indent
Reliable weather forecasting is of great importance in science, business, and society. The best performing data-driven models for weather prediction tasks rely on recurrent or convolutional neural networks, where some of which incorporate attention mechanisms. In this work, we introduce a novel model based on Transformer architecture for weather forecasting. The proposed Tensorial Encoder Transformer (TENT) model is equipped with tensorial attention and thus it exploits the spatiotemporal structure of weather data by processing it in multidimensional tensorial format. We show that compared to the classical encoder transformer, 3D convolutional neural networks, LSTM, and Convolutional LSTM, the proposed TENT model can better learn the underlying complex pattern of the weather data for the studied temperature prediction task. Experiments on two real-life weather datasets are performed. The datasets consist of historical measurements from weather stations in the USA, Canada and Europe. The first dataset contains hourly measurements of weather attributes for 30 cities in the USA and Canada from October 2012 to November 2017. The second dataset contains daily measurements of weather attributes of 18 cities across Europe from May 2005 to April 2020. Two attention scores are introduced based on the obtained tonsorial attention and are visualized in order to shed light on the decision-making process of our model and provide insight knowledge on the most important cities for the target cities. 


\end{abstract}


\end{frontmatter}


\section{Introduction}
In various application domains, one often encounters multivariate time series data, such as stock market prices, road traffic flows or weather conditions in different cities, to name a few \cite{lai2018modeling}. Predicting new trends and patterns in time series domains based on historical observations is the focus of many research studies. In particular, recent years have witnessed the development of advanced deep data-driven models and their promising results in weather forecasting applications \cite{MEHRKANOON_RepresLearning,trebing2020wind,fernandez2020deep,trebing2021smaat,gcn_t,aykas2021multistream,Ismail21,fernandez2021broad}.

Reliable weather forecasting is of great scientific, economic and social significance. Changing weather conditions impact many aspects of life, ranging from catastrophe and disaster management to many economic sectors including transport, agriculture, energy generation among others \cite{MEHRKANOON_RepresLearning}. The undergoing energy transformation to renewable energy production is only one of many sources of increasing demand for future weather conditions predication such as air temperature or wind speed \cite{kreuzer2020short}. High-quality temperature and wind forecasts increase the precision of estimates of future energy generation through wind and solar power \cite{kreuzer2020short}. Even a small increase in the prediction precision can have significant implications for the maximization of power generation \cite{ahmad2020maximizing} or the planning of energy distribution and power plant dispatching \cite{kreuzer2020short}.

In many fields such as physics, scientific computing, and deep machine learning, data appear in tensor (a multi-dimensional data structure) format. In particular, in the weather application domain, tensorial data consists of several weather variables measured at different time and space. Simply vectorizing tensor data loses useful structural information. Furthermore, higher-order correlations (correlations between the data dimensions) can only be discovered by simultaneously considering all the dimensions of the data. Tensorial machine learning based models to analyze high dimensional data have gained much attention in recent years \cite{dai2006tensor,nguyen2015tensor,ma2019tensorized}. 

It is the purpose of this paper to develop a Tensorized Encoder Transformer model that can preserve the spatiotemporal structure of the weather data and learn the input-output mapping by taking into account the underlying structure of the data. In particular, a new tensorial attention mechanism is designed that can provide additional insights on the explainability of the model by learning the attention weights using tensorial input data. In particular,  the learned attention weights can reveal the importance of the individual weather variables as well as the combination of weather variables and weather stations on the prediction of the target variables. This paper focuses on the temperature forecasting task. Three main contributions of our work are as follows: 
\begin{itemize}
    \item To develop a novel transformer architecture equipped with a new self-attention mechanism for analyzing 3D tensor input data. 
    \item To Introduce two attention scores based on the calculated tensorial attention. 
    \item To compare the proposed model with the classical encoder transformer, 3D convolutional neural network, LSTM and Convolutional LSTM models for the temperature prediction task. 
    \item To shed light on the explainability of the proposed model using data visualization techniques.
\end{itemize}

This paper is structured as follows. A brief overview of the related works for weather forecasting is given in section \ref{Related Works}. Our tensorial self-attention and transformer architecture with tensorial input are introduced in section \ref{Methodology}. The description of the used datasets, the obtained results and the analysis of the tensorial attention are reported in section \ref{Experiments}. The conclusions are drawn in section \ref{Conclusion}.

\section{Related Works}\label{Related Works}

\subsection{Weather elements forecasting}
Conventionally, weather forecasting is done by using Numerical Weather Prediction (NWP). NWP uses mathematical models to describe physical processes and weather conditions in the atmosphere or on the surface of the earth expressed by variables such as temperature, air pressure and wind \cite{MEHRKANOON_RepresLearning,marchuk2012numerical,richardson2007weather}. 
However, incomplete understanding of underlying complex atmospheric processes as well as the uncertainties in the initial conditions of the governed differential equation, may limit the accuracy of weather forecast \cite{MEHRKANOON_RepresLearning, soman2010review}. In addition, the computer simulation of NWP requires high computing power \cite{trebing2020wind, bauer2015quiet}.

In recent years, data-driven models have emerged which have proven to significantly reduce the processing time for weather forecasting \cite{trebing2020wind,ravuri2021skilful}. 
They rather rely on large amounts of historical weather observations that are used to extract and learn the underlying complex input-output mapping. The learned models are then used to predict future weather conditions \cite{MEHRKANOON_RepresLearning,chen2011comparison,kuligowski1998localized}.

However, the spatiotemporal nature of the underlying weather data along with its complex nonlinear behavior causes the prediction task to be nontrivial \cite{bartos2006nonlinear}. Thanks to the growing availability of weather data as well as advancements in computing power, many researchers have become motivated to explore a variety of deep data-driven models based on Artificial Neural Networks (ANN) \cite{salman2015weather}. In particular, Convolutional Neural Networks (CNN) and Recurrent Neural Networks (RNN) such as Long Short Term Memory (LSTM) neural networks have been successfully used to forecast hourly air temperature with significantly small errors for one time step ahead \cite{soman2010review, cifuentes2020air}. CNN-based models do not require any feature engineering; instead, they rely on feature extraction during the training of the network \cite{kreuzer2020short}. Mehrkanoon \cite{MEHRKANOON_RepresLearning} proposed a 3d-CNN based model for temperature forecasting. Klein et al. \cite{klein2015dynamic} introduced the Dynamic Convolutional Layer for short-range weather prediction task. In contrast to traditional CNN approaches, in Dynamic Convolutional Layer the filters vary from input to input during testing. RNNs, on the other hand, generally perform well when dealing with time-series data \cite{kreuzer2020short}. In particular, LSTM networks can better model long-term dependencies \cite{hochreiter1997long}. Long- and Short-term Time-series network (LSTNet) \cite{lai2018modeling} leverages both CNN and RNN to extract short-term local dependency patterns among variables and to discover long-term patterns for time series trends. The ConvLSTM network combines both CNN and LSTM architectures for precipitation prediction \cite{shi2015convolutional}. However, the sequential nature of processing elements in RNNs limits the parallelization during training, which becomes critical at longer sequence lengths \cite{vaswani2017attention}. 

Vaswani et al. \cite{vaswani2017attention} introduced an attention mechanism to model dependencies in sequences regardless of the distances between positions of the input elements. Often such attention mechanisms are combined with RNNs \cite{vaswani2017attention}. Temporal pattern attention LSTM (TPA-LSTM) \cite{shih2019temporal} utilizes an attention mechanism that allows the model not only to attend to relevant previous times but also to identify interdependencies among multiple features. The dual-stage attention-based recurrent neural network (DA-RNN) \cite{qin2017dual} incorporates a dual-stage attention scheme consisting of an encoder with an input attention mechanism and a decoder with a temporal attention mechanism. 

\subsection{Transformer architecture}
Transformer is an attention-based encoder-decoder architecture that was first introduced in 2017 \cite{vaswani2017attention}. It is designed to use the attention mechanism to handle sequential data by processing all input tokens at the same time. It replaces recurrent layers, commonly used in encoder-decoder architectures, with multi-headed self-attention where the dot-product attention is extended with a scaling factor of key-dimension $\frac{1}{\sqrt{d_k}}$ \cite{vaswani2017attention}. Consequently, the model does not rely on sequential processing of the data and therefore creates the possibilities for parallelization and reducing the training computational times \cite{vaswani2017attention}. Another commonly used attention mechanism is the additive attention, which uses a feed-forward neural network for the computation of the attention \cite{bahdanau2014neural}.

The Transformer is at the heart of many recent breakthroughs and developments in the field of sequence-to-sequence models. It has improved the state-of-the-art in many natural language processing tasks such as machine translation \cite{vaswani2017attention}, document generation \cite{liu2018generating} as well as syntactic parsing \cite{kitaev2018constituency}. The pretrained transformer language models, Bidirectional Encoder Representations from Transformers (BERT) and Generative Pre-trained Transformer (GPT) have already proved their effectiveness in tasks such as natural language understanding and question answering \cite{devlin2018bert,radford2018improving}. Generative models such as the Image Transformer \cite{parmar2018image} and Music Transformer \cite{huang2018music} have shown their capability to generate convincingly natural images and musical compositions based on human evaluation.

Shen et al. \cite{shen2018tensorized} developed Multi-mask Tensorized Self-Attention (MTSA), which captures the dependency between every two tokens using the scaled dot-product self-attention and utilizes the multi-dim self-attention mechanism \cite{shen2018disan} to estimate the contribution of each token to the given NLP task on each feature dimension. Zhou et al. \cite{zhou2021informer} proposed the Informer architecture, which can handle extremely long sequences efficiently with its ProbSparse attention mechanism, which generates different sparse query-key pairs for each head to avoid information loss.

Ma et al. \cite{ma2019tensorized} used a tensorized transformer to compress model parameters and increase performance. In particular, \cite{ma2019tensorized} introduced Multi-linear attention for language modeling approaches, which uses Tucker decomposition \cite{tucker1966some, li2017bt} in Single-block attention to reconstruct the scaled dot-product attention and Block-Term Tensor Decomposition \cite{de2008decompositions} for the multi-head attention mechanism. This paper proposes a Tensorized Encoder Transformer (TENT) for temperature forecasting task. The model is equipped with a new tensorial attention mechanism. As opposed to \cite{ma2019tensorized}, our proposed TENT model receives 3-dimensional tensors as input and maintains the tensorial format in the calculation of the self-attention layer. Thanks to the introduced tensorial self-attention mechanism, TENT model is able to capture the dependencies between time steps (lags) and the weather stations by simultaneously considering all the dimensions of the data. In the subsequent sections, the proposed model will be explained in detail.

\section{Proposed Methodology}\label{Methodology}

\subsection{Tensorized Encoder Transformer}
\begin{figure}[!ht]
  \centering
  \includegraphics[width=1\linewidth]{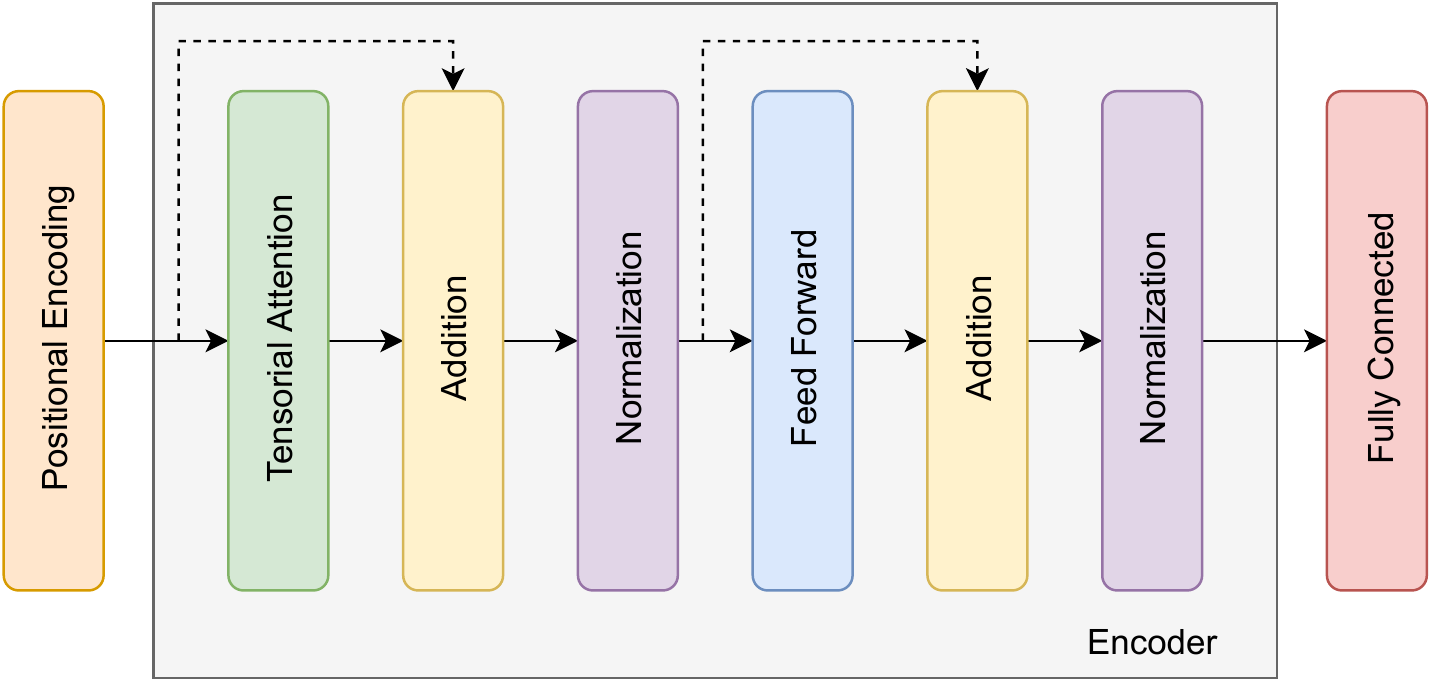}
  \caption{Model architecture of Tensorized Encoder Transformer (TENT).}
  \label{fig:model}
\end{figure}

Here, we extend the original Transformer \cite{vaswani2017attention} and introduce the Tensorized Encoder Transformer (TENT) model suitable for learning complex underlying patterns of tensorial input data. In particular, the model processes weather inputs of 3D tensorial format $X \in{ \R^{T \times C \times F}}$, where $T$ is the number of time steps, $C$ and $F$ represent 2D features. 2D features of size $C \times F$ represent a matrix of weather variables such as temperature, humidity, and wind speed measured at different geographical locations. More precisely, different cities are located along dimension $C$ and the weather variables are along dimension $F$. 
The proposed model is an encoder-only Transformer architecture that consists of a positional encoding layer, encoder layer followed by a fully-connected layer with linear activation as shown in Fig. \ref{fig:model}. The encoder layer uses the tensorial attention that will be introduced in sections \ref{Tensorial self-attention} and \ref{Tensorial multihead-attention}. Subsequent to the tensorial attention are a residual connection and a normalization layer. Here a fully connected feed-forward network with two linear transformations and ReLU activations is used after the tensorial attention layer. In line with \cite{vaswani2017attention}, it is then followed by a second residual connection and normalization layer.

\subsection{Positional Encoding}
We use a fixed positional encoding to equip the model with time sequence information. The positional encoding \cite{vaswani2017attention} is calculated for the time step $T$ and city $C$ axes according to Eq. (\ref{eq:1}), where, $pos$ is the position in time step axis and $i$ is the position in the city axis. We broadcast the obtained values along the weather variables axis $F$. Hence, the values of the positional encoding are identical across the $F$ axis and differ in $T$ and $C$ axes.

\begin{equation} \label{eq:1}
\begin{aligned}
\begin{cases}
PE_{(pos,2i)} = \sin(pos/10000^{2i/C}), \\
PE_{(pos,2i+1)}= \cos(pos/10000^{2i/C}).
\end{cases}
\end{aligned}
\end{equation}

\subsection{Tensorial self-attention} \label{Tensorial self-attention}
\begin{figure}[!ht]
  \centering
  \includegraphics[width=1\linewidth]{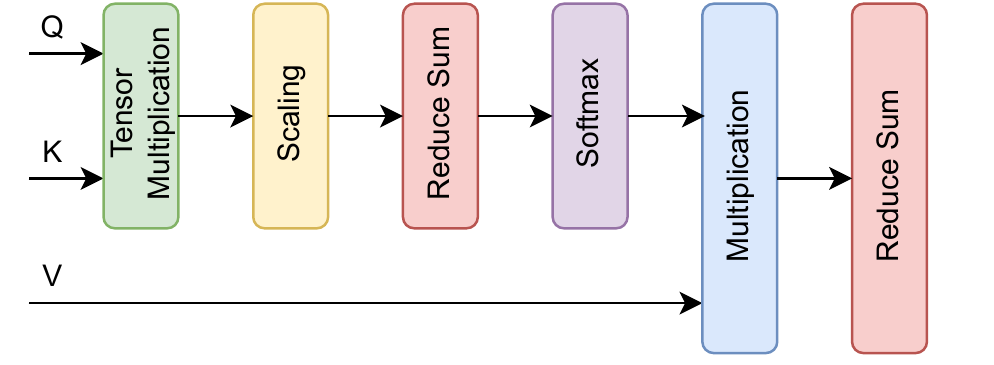}
  \caption{Tensorial self-attention.}
  \label{fig:self-attention}
\end{figure}

To formulate tensorial self-attention, see Fig. \ref{fig:self-attention}, the following slice notation is used. For a tensor $A \in \R^{X \times Y \times Z}$ the $x$ slice is the matrix $(A_{y,z})_x \in \R^{Y \times Z}$ formed from all the values of the tensor $A$ where the first dimension is set to $x$. Similarly, the $x, y$ slice of $A$ is the vector $(A_z)_{x, y} \in \R^{Z}$ formed from the values of $A$ where the first and second dimensions are set to $x$ and $y$ respectively. We denote the dimension of the slice by the small letter corresponding to the particular dimension. 

\begin{figure}[!ht]
  \centering
  \includegraphics[width=.7\linewidth]{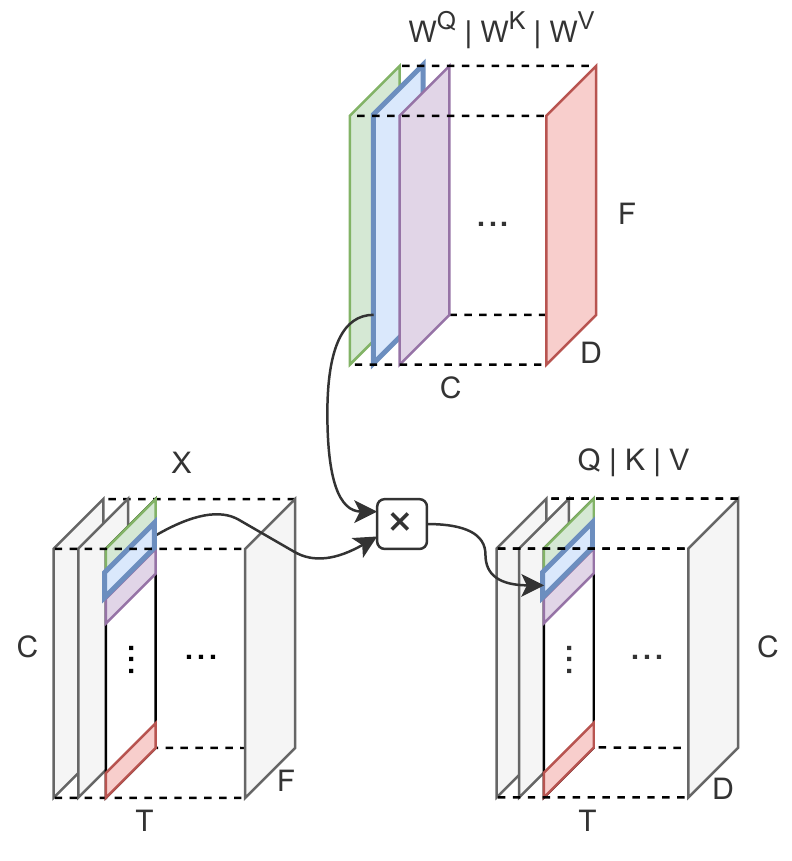}
  \caption{The visualization of the slice multiplication for obtaining query, key and value tensors from the tensor input $X$ and weight tensors $W^Q$, $W^K$ and $W^V$. In this figure, vector slices in the input are multiplied by the matrix slices of weights of the same color to form a vector slice of that color in the output tensors $Q$, $K$, and $V$.}
  \label{fig:slice-multiplication}
\end{figure}

The input of the tensorial self-attention is a 3D tensor $X \in \R^{T \times C \times F}$. In the first step of the tensorial self-attention, the 3D Query ($Q$), Key ($K$) and Value ($V$) tensors which all have the same dimension, i.e. $\R^{T \times C \times D}$, are calculated by multiplying the input tensor $X$ with weight tensors. Here $D$ is obtained by dividing the key-dimension $d_k$ by the number of heads, which are both hyper-parameters. More precisely, three separate 3D weight tensors $W^Q, W^K$ and $W^V \in \R ^ {C \times F \times D}$ are used. Each $t, c$ slice of $Q$, $K$ and $V$ (i.e. $(Q_d)_{t,c}$, $(K_d)_{t,c}$ and $(V_d)_{t,c}$ respectively) is calculated by multiplying the $t, c$ slice of $X$, denoted by $(X_f)_{t,c}$, and the $c$ slice of $W^Q$, $W^K$ and $W^V$ (i.e. $(W^Q_{f,d})_c$, $(W^K_{f,d})_c$ and $(W^V_{f,d})_c$ respectively) as 
follows:

\begin{equation} \label{eq:3}
\begin{aligned} 
\begin{cases}
&(Q_d)_{t,c} = (X_f)_{t,c} \times (W^Q_{f,d})_c \;,\; \forall t=1,\dots,T, \;\; c=1,\dots,C,\\ 
&(K_d)_{t,c} = (X_f)_{t,c} \times (W^K_{f,d})_c \;,\; \forall t=1,\dots,T, \;\; c=1,\dots,C,\\
&(V_d)_{t,c} = (X_f)_{t,c} \times (W^V_{f,d})_c  \;,\; \forall t=1,\dots,T, \;\; c=1,\dots,C.
\end{cases}
\end{aligned}
\end{equation}
Fig. \ref{fig:slice-multiplication} illustrates the visualization of the multiplication operations of Eq. (\ref{eq:3}).

In the next step, each time step $t$ slice of $Q$ (denoted by $(Q_{c,d})_t$) is multiplied with the transpose of each time step $t'$ slice of $K$ (denoted by $((K_{c',d})_{t'})^T$).

These multiplications, see Eq. (\ref{eq:6a}), result in matrices $(\tilde{R}_{c,c'})_{t,t'}$ of shape $C \times C'$, which are slices of a tensor $\tilde{R} \in \R^{T \times T' \times C \times C'}$, where $T$ and $C$ are the first and second dimensions of $Q$ tensor and $T'$ and $C'$ are the first and second dimensions of $K$ tensor. Following the lines of \cite{vaswani2017attention}, in Eq. (\ref{eq:6b}) the tensor $\tilde{R}$ is summed over the last dimension and divided element-wise by the square root of $D$ resulting in a tensor $R \in \R^{T \times T' \times C}$. 
\begin{subequations} \label{eq:6}
\begin{align}
(\tilde{R}_{c,c'})_{t,t'} &= (Q_{c,d})_{t} \times ((K_{c',d})_{t'})^T \;,\; \forall t,t'=1,\dots,T,  \label{eq:6a} \\ 
R &= \frac{1}{\sqrt{D}} \sum_{c'=1}^{C} (\tilde{R}_{t,t',c})_{c'}. \label{eq:6b}
\end{align}
\end{subequations}
Next, we apply a softmax function over each $t,t'$ slice of $R$, producing the attention tensor $S \in \R^{T \times T' \times C}$ as follows:
\begin{equation} \label{eq:7}
(S_{c})_{t,t'} = \text{softmax}((R_{c})_{t,t'}) \;\;\; \forall t,t'=1,\dots,T.
\end{equation}
Lastly, the output of the tensorial self-attention, i.e. $Z \in \R^{T \times C \times D}$, is calculated by taking the Hadamard product from the $t, t'$ slice of the attention weights $(S_c)_{t,t'}$ and the $t'$ slice of $V$ (i.e. $(V_{c,d})_{t'}$), and summing over the second dimension (corresponding to $t'$). One should note that $(S_c)_{t,t'}$ slice is first broadcasted to shape $C \times D$ to match the shape of $(V_{c,d})_{t'}$ slice. This calculation can be seen in Eq. (\ref{eq:8}).

\begin{equation} \label{eq:8}
(Z_{c,d})_{t} = \sum_{t'=1}^{T} \textrm{broadcast}((S_c)_{t,t'}) \circ (V_{c,d})_{t'} \;\;\; \forall t=1,\dots,T.
\end{equation}

\subsection{Tensorial multi-head attention} \label{Tensorial multihead-attention}
\begin{figure}[!ht]
  \centering
  \includegraphics[width=.85\linewidth]{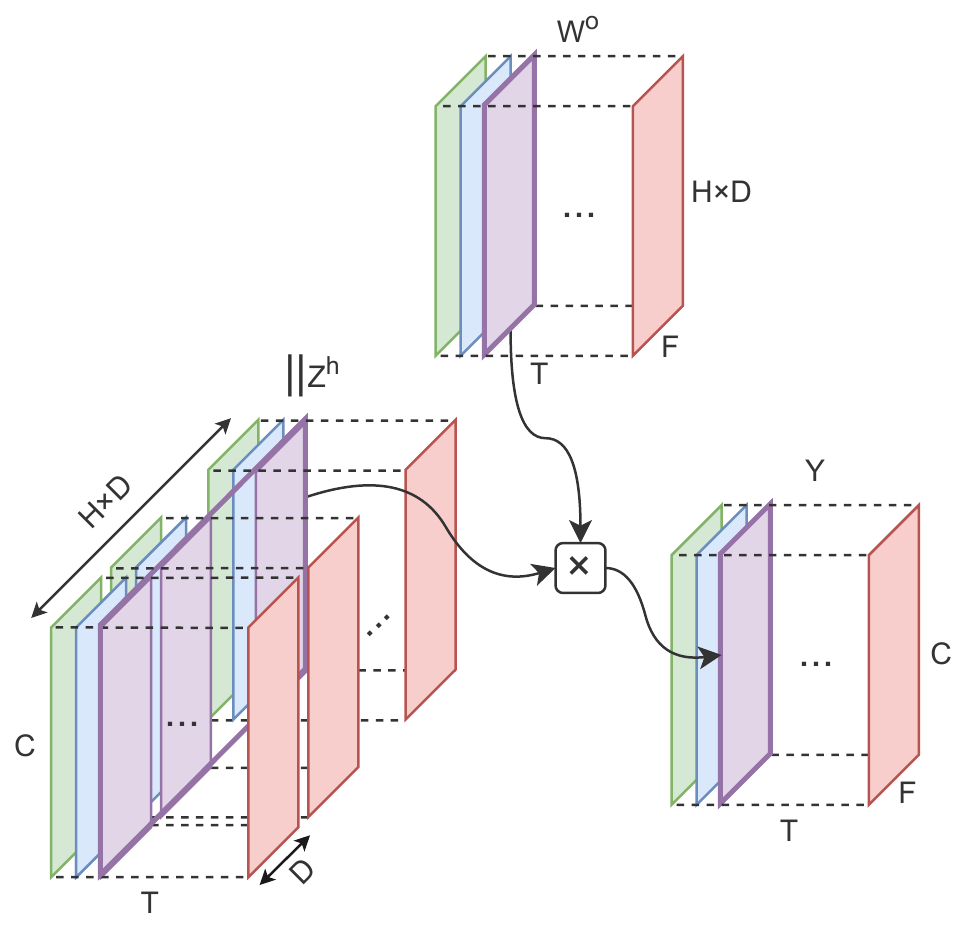}
  \caption{The visualization of the slice multiplication for obtaining the output of the multi-head attention $Y$ from the concatenated self-attention output tensors $\concat Z^h$ and weight tensor $W^o$. In this figure, time slices in the concatenated self-attention output tensors are multiplied by the time slices of weight tensor of the same color to form a time slice of that color in the output $Y$.}
  \label{fig:multihead-attention}
\end{figure}

Here, we calculate tensorial self-attention $H$ times, which leads to tensorial multi-head attention.
We extend the calculation of the multi-head attention \cite{vaswani2017attention} to tensorial inputs. Each of these heads correspond to the output of the tensorial self-attention $Z^h$, where $h$ indicates the index of the head. The outputs are concatenated along the last dimension $\concat_{h=1}^H Z^h$, resulting in a $T \times C \times (H \times D)$ tensor. Next the output tensor of multi-head attention $Y \in \R^{T \times C \times F}$ is obtained by multiplying each time slice, $( (\concat_{h=1}^H Z^h)_{c,h\cdot d})_t$, with a time slice of a weight tensor $W^o$ with shape $T \times (H \times D) \times F$ as follows:
\begin{equation} \label{eq:9}
\begin{aligned}
& (Y_{c,f})_{t} = ( (\concat_{h=1}^H Z^h)_{c,h\times d})_t \times (W^o_{h\times d,f})_t 
& \forall t=1,\dots,T,
\end{aligned}
\end{equation}
The visualization of the multiplication operation of Eq. (\ref{eq:9}) is shown in Fig. \ref{fig:multihead-attention}.

\subsection{Attention Aggregation} \label{Attention aggregation}
The attention weights have been previously used for feature selection purposes as well as model explainability \cite{wiegreffe2019attention, Gui_Ge_Hu_2019}. Here, the attention tensors $S$ (see Eq. (\ref{eq:7})) are aggregated to shed light on the model predictions. Let us denote the extracted attention tensors by $S^h \in{ \R^{T \times T' \times C}}$, where $h$ is the head index, $T$ and $T'$ are the numbers of time steps due to the Query and the Key respectively and $C$ is the number of cities. In order to show the correlation between heads and cities, attention scores ($AS_{c}^{h}$) are computed as follows: 
\begin{equation} \label{eq:s_hc}
AS_{c}^{h}=\sum_{t=1}^{T}\sum_{t'=1}^{T'} S_{t,t',c}^h \;,\;\; \forall h=1,\dots,H, \;\;\; c=1,\dots,C.
\end{equation}
Furthermore, in order to show the contribution of each city to the prediction, we calculate the attention scores $AS_c$ for each city as follows: 
\begin{equation} \label{eq:s_c}
AS_{c}=\sum_{h=1}^{H} AS_{c}^{h} \;,\;\; \forall c=1,\dots,C.
\end{equation}

\begin{table*}[!ht]
\setlength{\tabcolsep}{4pt}
\renewcommand{\arraystretch}{1.0}
\centering
\caption{Hyper-parameters used for all the models for USA-Canada as well as Europe datasets.}
\label{tab:hyper-parameters}
\begin{tabular}{l|cc|cc|cc|cc|cc}
\multirow{2}{*}{Hyper-parameter}    & \multicolumn{2}{c|}{TENT} & \multicolumn{2}{c|}{Transformer} & \multicolumn{2}{c|}{3d CNN} & \multicolumn{2}{c|}{LSTM} & \multicolumn{2}{c}{ConvLSTM} \\
    & \scriptsize{USA-Canada} & \scriptsize{Europe} & \scriptsize{USA-Canada} & \scriptsize{Europe} & \scriptsize{USA-Canada} & \scriptsize{Europe} & \scriptsize{USA-Canada} & \scriptsize{Europe} & \scriptsize{USA-Canada} & \scriptsize{Europe} \\  \hline 
Layer Number & 1 & 1 & \multicolumn{2}{c|}{1} & \multicolumn{2}{c|}{-} & 1 & 1 & 1 & 4 \\
Head Number & 8 & 1 & \multicolumn{2}{c|}{4} & \multicolumn{2}{c|}{-} & - & - & - & - \\
Key-Dimension & 16 & 1 & \multicolumn{2}{c|}{16} & \multicolumn{2}{c|}{-} & - & - & - & - \\
Dense Units & 32 & 6 & \multicolumn{2}{c|}{32} & \multicolumn{2}{c|}{128} & - & - & - & - \\
Filters & - & - & \multicolumn{2}{c|}{-} & \multicolumn{2}{c|}{10} & - & - & 8 & 16 \\
Kernel Size & - & - & \multicolumn{2}{c|}{-} & \multicolumn{2}{c|}{2} & - & - & 11 & 3 \\
Hidden Units & - & - & \multicolumn{2}{c|}{-} & \multicolumn{2}{c|}{-} & 64 & 128 & - & - \\
Learning Rate & Schedule & Schedule & \multicolumn{2}{c|}{Schedule} & \multicolumn{2}{c|}{$10^{-4}$} & $10^{-4}$ & $10^{-2}$ & $10^{-4}$ & $10^{-4}$ \\
Batch Size & 96 & 128 & \multicolumn{2}{c|}{128} & \multicolumn{2}{c|}{128} & 256 & 512 & 128 & 128 \\
\end{tabular}
\end{table*}

\begin{figure*}[!ht]
\center{}
  \begin{subfigure}{.4\linewidth}
    \figuretitle{Average over time steps}
    \includegraphics[width=\linewidth]{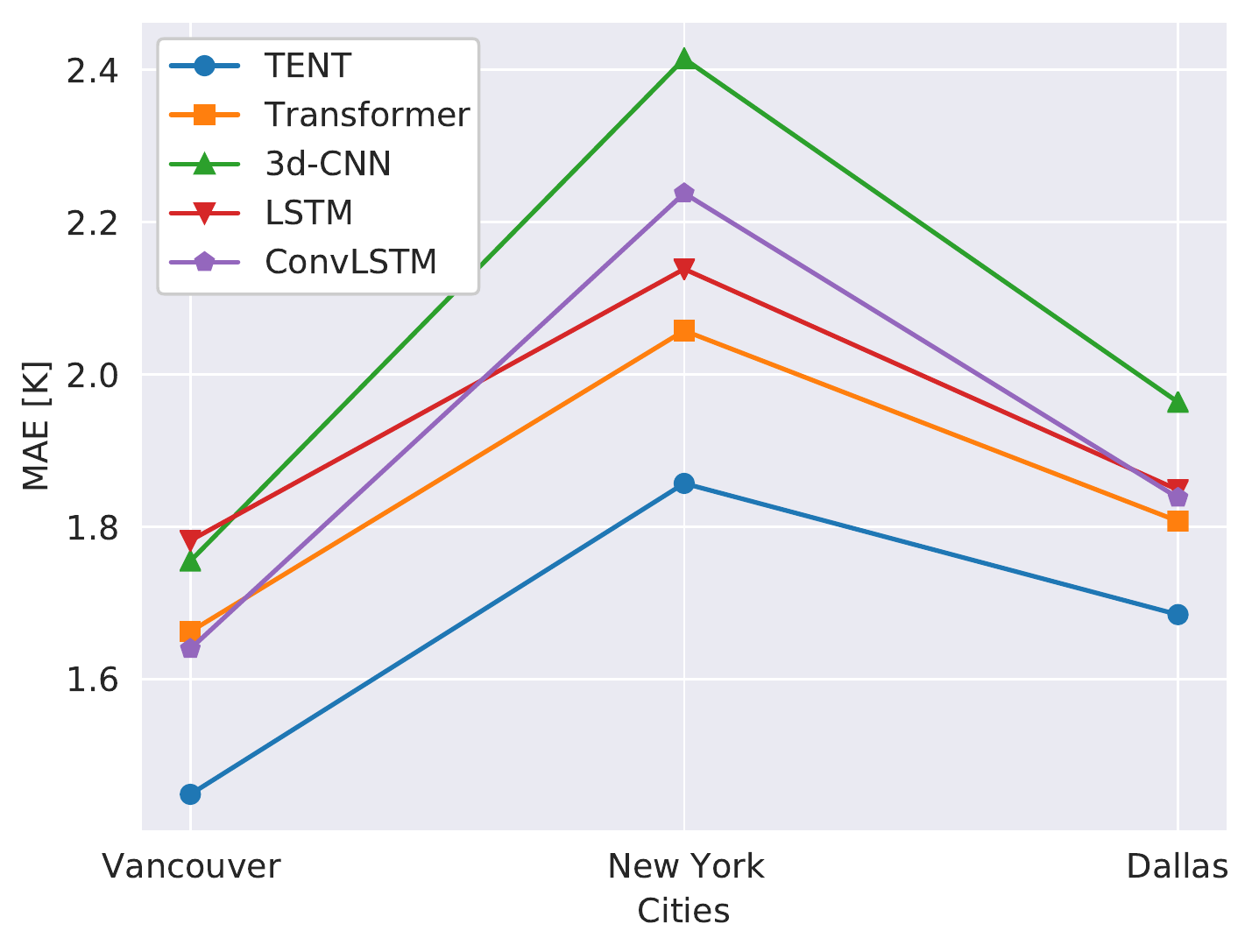}
    \caption{}
    \label{fig:mae_usa_aggregated_cities}
  \end{subfigure}\hspace{0.05\textwidth}
  \begin{subfigure}{.4\linewidth}
    \figuretitle{Average over cities}
    \includegraphics[width=\linewidth]{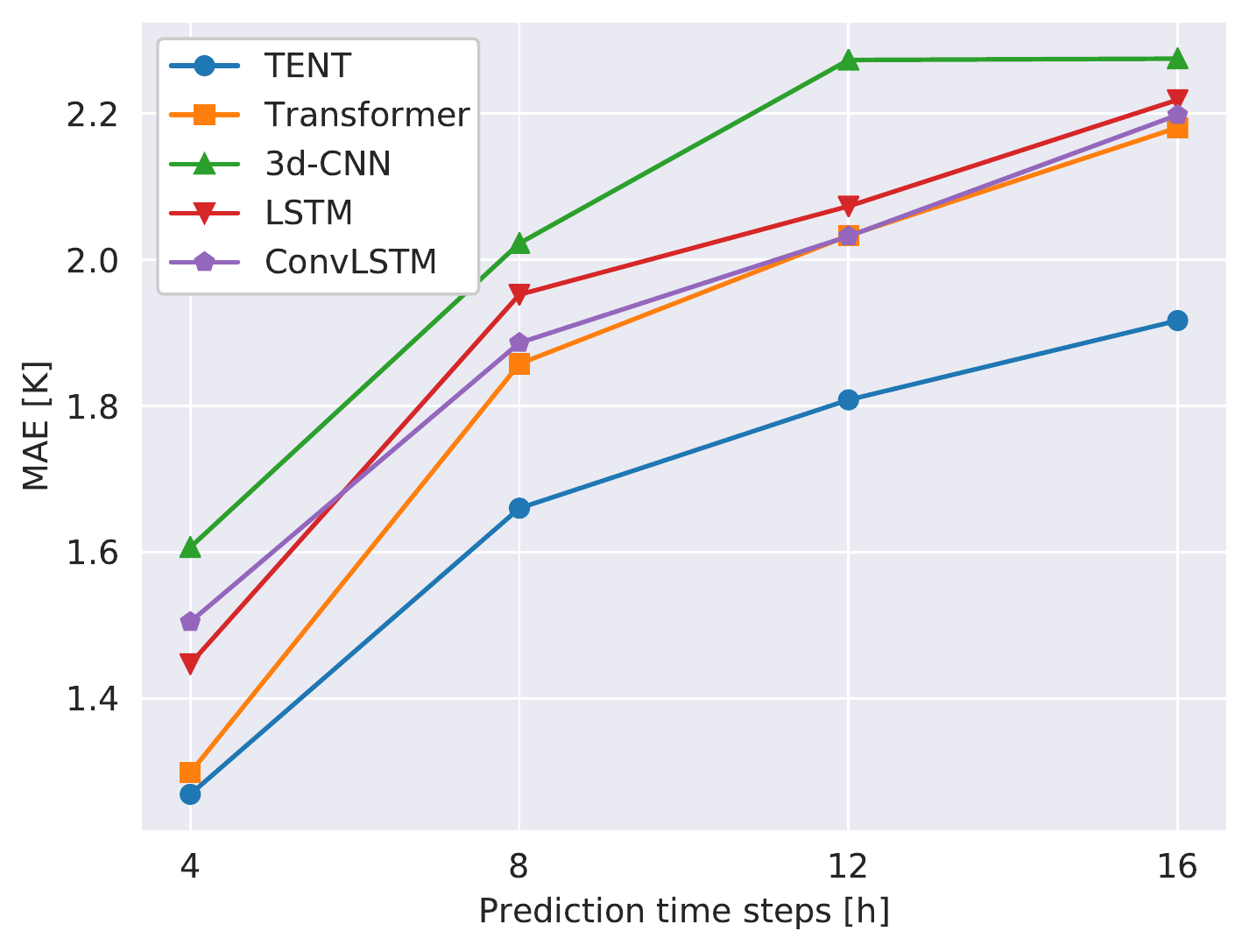}
    \caption{}
    \label{fig:mae_usa_aggregated_times}
  \end{subfigure}
  
  \caption{The obtained test MAE of the models for the \textbf{USA and Canada} datasets averaged over cities (\ref{fig:mae_usa_aggregated_cities}) and prediction time steps (\ref{fig:mae_usa_aggregated_times}).}
  \label{fig:mae_usa_aggregated}
\end{figure*}

\section{Experiments}\label{Experiments}
We apply our proposed TENT model on two datasets depicted in Table \ref{tab:dataset} and compare the obtained results with four other models, i.e. classical encoder Transformer \cite{vaswani2017attention} (with the input tensor flattened), 3d-CNN \cite{MEHRKANOON_RepresLearning}, LSTM \cite{LSTM} and Convolutional LSTM \cite{ConvLSTM}. Following the lines of \cite{MEHRKANOON_RepresLearning,trebing2020wind}, the Mean Absolute Error (MAE) and Mean Squared Error (MSE) metrics are used to evaluate the performance of the models as follows:

\begin{equation} \label{eq:10}
\text{MAE}=\frac{\sum_{i=1}^{n} \lvert y_i-\hat{y}_i\rvert}{n},\;\;\;\;\; \text{MSE}=\frac{\sum_{i=1}^{n} (y_i-\hat{y}_i)^2}{n}.
\end{equation}

\begin{table}[!ht]
\setlength{\tabcolsep}{4pt}
\renewcommand{\arraystretch}{1.2}
\centering
\caption{Size of train, validation and test datasets and number of cities and features for USA-Canada and Europe datasets.}
\label{tab:dataset}
\begin{tabular}{c|ccc|c|c}
           & \multicolumn{3}{c|}{Size} & \multirow{2}{*}{Cities} & \multirow{2}{*}{Features} \\
Dataset    & Train & Validation & Test &                         &                           \\ \hline
USA-Canada & 35362 & 1024       & 7997 & 30                      & 11                        \\
Europe     & 3854  & 512        & 1086 & 18                      & 19                       
\end{tabular}
\end{table}

Here, $n$ is the number of samples, $\hat{y}_i$ is the model prediction and $y_i$ is the true measurement. The training is done with MSE loss function for a maximum of $300$ epochs and early stopping based on validation loss with patience $20$ is used. Adam optimizer \cite{kingma2014adam} is employed for all models. Both TENT and Transformer use custom learning rate schedule \cite{vaswani2017attention} while 3d-CNN, LSTM and Convolutional LSTM use a learning rate adapted by Adam optimizer. The empirically found optimal hyper-parameters for each model are tabulated in Table \ref{tab:hyper-parameters}.

\begin{table*}[hbt!]
\centering
\caption{The obtained test MAE and MSE of the \textbf{USA-Canada} dataset for temperature prediction.}
\label{tab:usa_results}
\begin{tabular}{ll|rrrr|rrrr}
              &             & \multicolumn{4}{c|}{MAE}                                                                                                 & \multicolumn{4}{c}{MSE}                                                                                                  \\  
       Station                    &         Model    & \multicolumn{1}{l}{4 hours} & \multicolumn{1}{l}{8 hours} & \multicolumn{1}{l}{12 hours} & \multicolumn{1}{l|}{16 hours} & \multicolumn{1}{l}{4 hours} & \multicolumn{1}{l}{8 hours} & \multicolumn{1}{l}{12 hours} & \multicolumn{1}{l}{16 hours}  \\ 
\hline
\multirow{5}{*}{Vancouver} & TENT        & \boldmath{$1.191$}            & \boldmath{$1.487$}            & \boldmath{$1.546$}             & \boldmath{$1.571$}                       & \boldmath{$2.383$}            & \boldmath{$3.733$}            & \boldmath{$4.026$}             & \boldmath{$4.116$}                       \\
                           & Transformer & $1.231$                     & $1.672$                     & $1.841$                      & $1.907$                       & $2.592$                     & $4.903$                     & $5.809$                      & $6.127$                       \\
                           & 3d CNN      & $1.486$                     & $1.855$                     & $1.856$                      & $1.823$              & $3.650$                     & $5.688$                     & $5.744$                      & $5.520$              \\ 
                           & LSTM      & $1.436$                     & $1.897$                     & $1.856$                      & $1.939$              & $3.399$                     & $6.043$                     & $5.746$                      & $6.378$              \\ 
                           & ConvLSTM      & $1.482$                     & $1.696$                     & $1.685$                      & $1.696$              & $3.656$                     & $4.695$                     & $4.582$                      & $4.697$              \\ 
\hline
\multirow{5}{*}{New York}  & TENT         & \boldmath{$1.231$} & \boldmath{$1.664$} & \boldmath{$1.860$} & \boldmath{$1.984$} & \boldmath{$2.538$} & \boldmath{$4.604$} & \boldmath{$5.779$} & \boldmath{$6.458$}              \\
                           & Transformer  & $1.238$ & $1.858$ & $1.987$ & $2.146$ & $2.566$ & $5.787$ & $6.617$ & $7.748$                       \\
                           & 3d CNN      & $1.499$                     & $1.896$                     & $2.131$                      & $2.329$                       & $3.704$                     & $5.950$                     & $7.455$                      & $8.879$                       \\ 
                           & LSTM       & $1.311$ & $1.834$ & $2.039$ & $2.210$ & $2.917$ & $5.712$ & $6.970$ & $8.237$              \\ 
                           & ConvLSTM       & $1.338$ & $1.829$ & $1.992$ & $2.194$ & $2.967$ & $5.553$ & $6.571$ & $7.990$              \\ 
\hline
\multirow{5}{*}{Dallas}    & TENT         & \boldmath{$1.383$} & \boldmath{$1.830$} & \boldmath{$2.020$} & \boldmath{$2.195$} & \boldmath{$3.380$} & \boldmath{$5.796$} & \boldmath{$6.977$} & \boldmath{$8.293$}              \\
                           & Transformer  & $1.426$ & $2.043$ & $2.271$ & $2.489$ & $3.836$ & $7.533$ & $9.268$ & $10.97$                       \\
                           & 3d CNN      & $1.835$                     & $2.316$                     & $2.833$                      & $2.673$                       & $5.587$                     & $9.159$                     & $13.46$                      & $11.96$          \\ 
                           & LSTM       & $1.596$ & $2.126$ & $2.325$ & $2.507$ & $4.724$ & $8.403$ & $9.749$ & $10.98$              \\ 
                           & ConvLSTM       & $1.694$ & $2.134$ & $2.419$ & $2.704$ & $4.949$ & $7.790$ & $9.757$ & $12.34$              \\ 
                           \hline            
\end{tabular}
\end{table*}

\begin{figure*}[hbt!]
\center{}
  \begin{subfigure}{0.48\linewidth}
    \figuretitle{Vancouver - 4 hours in the future}
    \includegraphics[width=\linewidth]{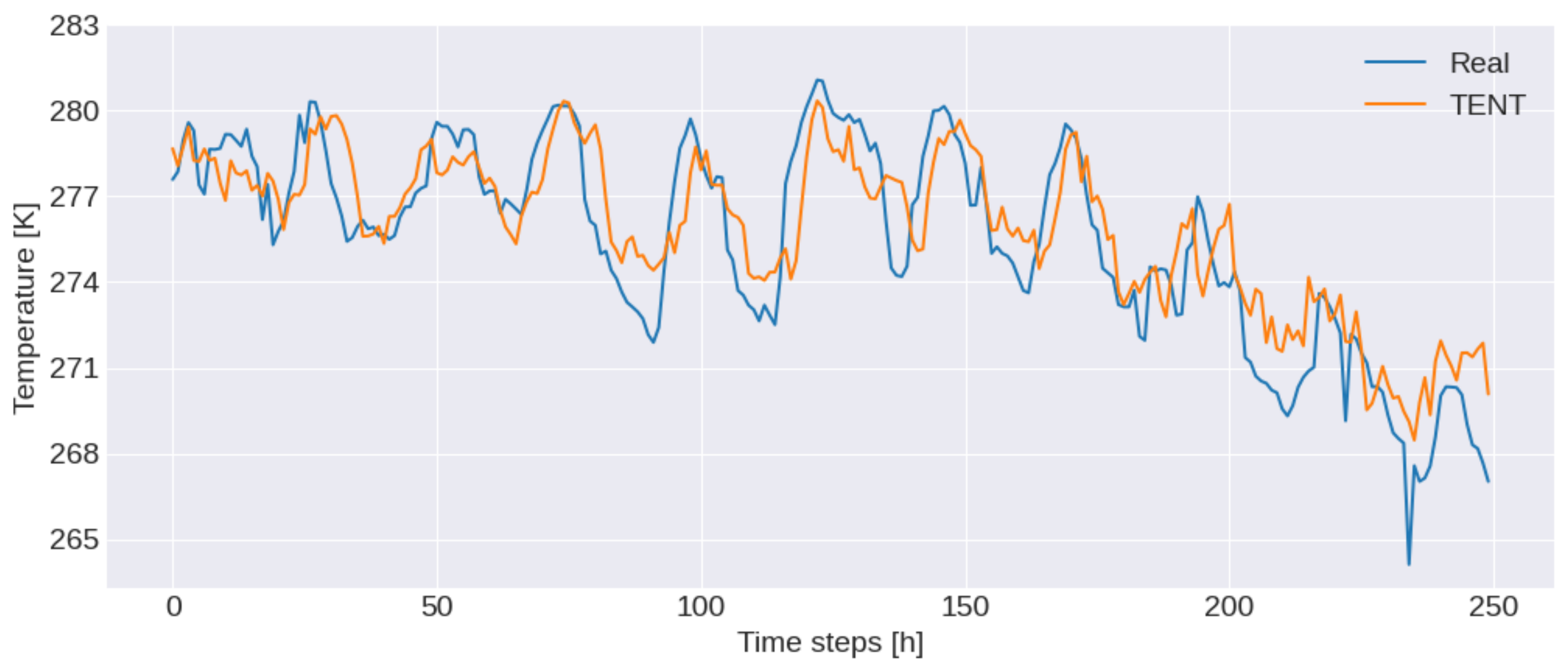}\vspace{-0.5em}
    \caption{}\vspace{1em}
    \label{fig:PredRealEUA_lag4}
  \end{subfigure}\hfill
  \begin{subfigure}{0.48\linewidth}
    \figuretitle{Vancouver - 8 hours in the future}
    \includegraphics[width=\linewidth]{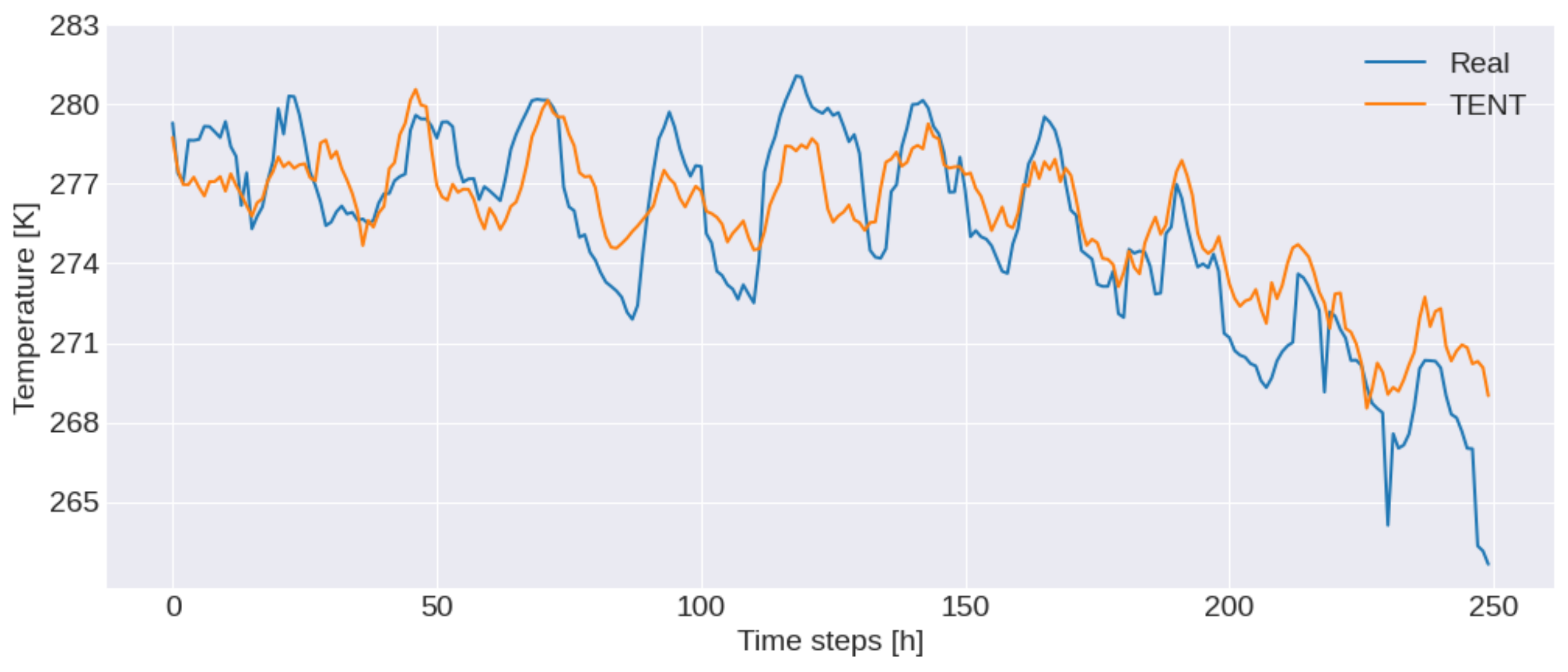}\vspace{-0.5em}
    \caption{}\vspace{1em}
    \label{fig:PredRealEUA_lag8}
  \end{subfigure}\hfill
  \begin{subfigure}{0.48\linewidth}
    \figuretitle{Vancouver - 12 hours in the future}
    \includegraphics[width=\linewidth]{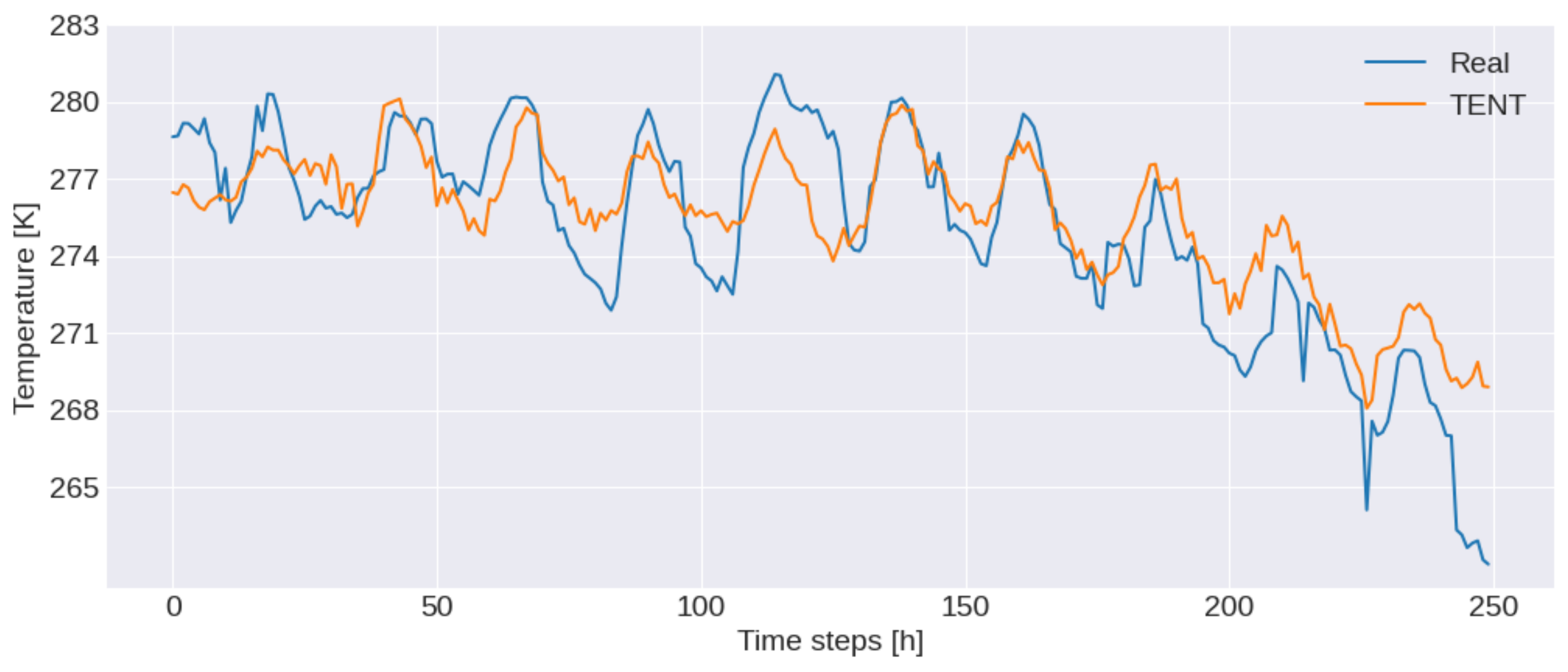}\vspace{-0.5em}
    \caption{}
    \label{fig:PredRealEUA_lag12}
  \end{subfigure}\hfill
  \begin{subfigure}{0.48\linewidth}
    \figuretitle{Vancouver - 16 hours in the future}
    \includegraphics[width=\linewidth]{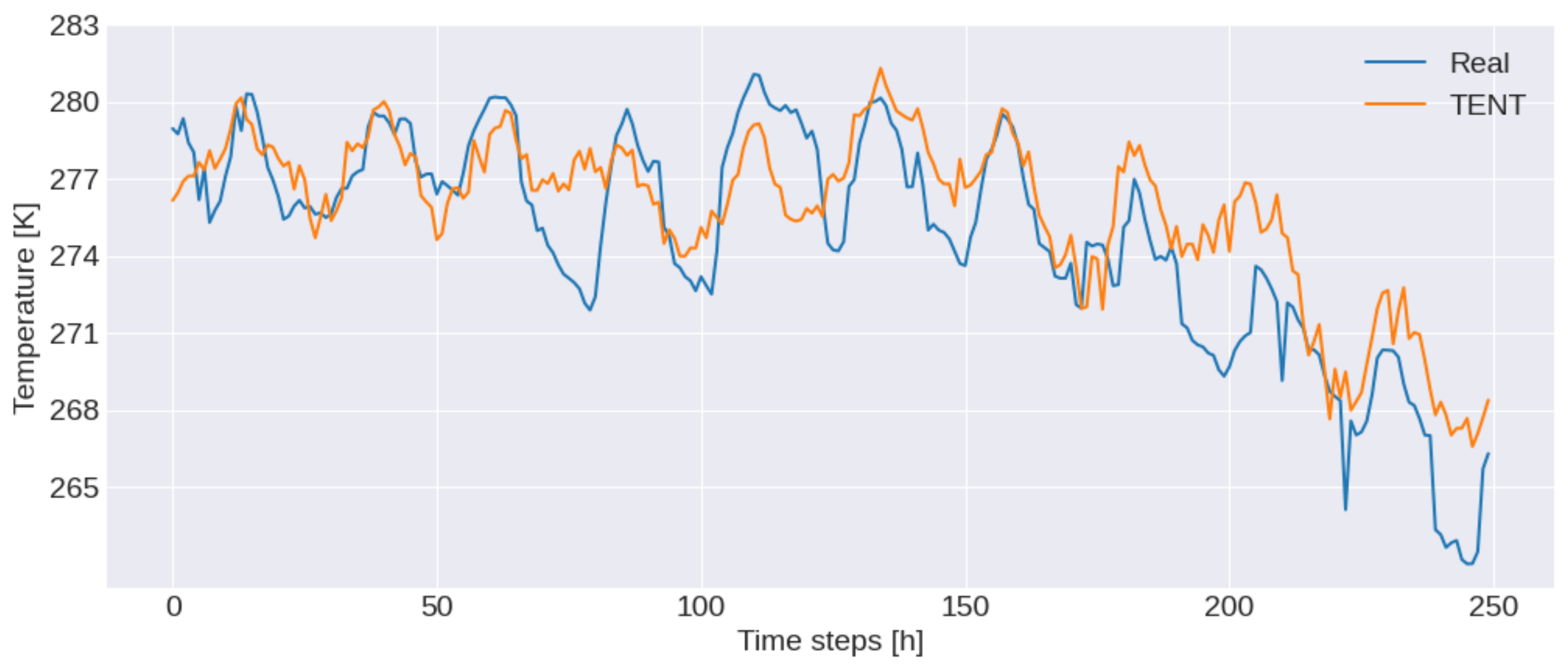}\vspace{-0.5em}
    \caption{}
    \label{fig:PredRealEUA_lag16}
  \end{subfigure}\hfill
  \caption{The comparison between the predictions of TENT model and the real measurements for \textbf{hourly temperature} of the test set of \textbf{Vancouver}.}
  \label{fig:PredRealEUA}
\end{figure*}

\subsection{USA-Canada Dataset}
This dataset contains hourly measurements of the weather attributes such as humidity, air pressure, temperature, weather description, wind direction and wind speed for 30 cities in the USA and Canada from October 2012 to November 2017. 

\begin{table*}[hbt!]
\centering
\caption{The obtained test MAE and MSE of the \textbf{Europe} dataset for temperature prediction.}
\label{tab:eu_results}
\begin{tabular}{ll|rrr|rrr}
                            &             & \multicolumn{3}{c|}{MAE}                                                              & \multicolumn{3}{c}{MSE}                                                               \\
        Station                    &    Model         & \multicolumn{1}{l}{2 days} & \multicolumn{1}{l}{4 days} & \multicolumn{1}{l|}{6 days} & \multicolumn{1}{l}{2 days} & \multicolumn{1}{l}{4 days} & \multicolumn{1}{l}{6 days}  \\ 
\hline
\multirow{5}{*}{Barcelona}  & TENT        & $2.327$                    & $2.807$                    & $2.947$                     & $9.533$                    & $13.79$                    & $15.14$                     \\
                            & Transformer & $2.608$                    & $2.901$                    & $3.047$                     & $11.70$                    & $14.66$                    & $15.92$                     \\
                            & 3d CNN      & $2.502$                    & $3.015$                    & $3.059$                     & $10.73$                    & $14.64$                    & $15.74$                     \\ 
                            & LSTM        & \boldmath{$2.303$}         & $2.801$          & \boldmath{$2.931$}           & \boldmath{$9.354$}                   & $13.32$                    & $14.93$                     \\ 
                            & ConvLSTM    & $2.759$          & \boldmath{$2.787$}                   & $2.948$                     & $12.82$          & \boldmath{$13.22$}         & \boldmath{$14.90$}                    \\ 
\hline
\multirow{5}{*}{Maastricht} & TENT        & $4.164$          & \boldmath{$4.900$}         & \boldmath{$5.140$}                    & $28.45$          & \boldmath{$38.50$}         & \boldmath{$42.70$}                    \\
                            & Transformer & $4.370$                    & $5.239$                    & $5.649$                     & $30.89$                    & $43.23$                    & $50.67$                     \\
                            & 3d CNN      & $4.276$                    & $5.078$                    & $5.609$                     & $28.82$                    & $40.51$                    & $49.41$                     \\ 
                            & LSTM        & \boldmath{$3.982$}         & $5.036$                    & $5.373$           & \boldmath{$24.86$}                   & $39.48$                    & $46.59$                     \\ 
                            & ConvLSTM    & $4.578$                    & $5.063$                    & $5.222$                     & $32.69$                    & $39.89$                    & $43.28$                     \\ 
\hline
\multirow{5}{*}{Munich}     & TENT        & $3.696$                    & $4.835$                    & $5.400$                     & $21.97$                    & $36.04$                    & $45.30$                     \\
                            & Transformer & $3.836$                    & $4.986$                    & $5.275$                     & $23.94$                    & $39.05$                    & $43.56$                     \\
                            & 3d CNN      & $3.931$                    & $5.049$                    & $5.262$                     & $24.87$                    & $39.57$                    & $43.50$                     \\
                            & LSTM        & \boldmath{$3.551$}         & \boldmath{$4.730$}         & $5.189$           & \boldmath{$20.25$}         & \boldmath{$34.01$}                   & $42.73$                     \\ 
                            & ConvLSTM    & $3.794$                    & $4.830$                    & \boldmath{$5.023$}          & $22.44$                    & $35.01$          & \boldmath{$37.76$}                    \\ 
                            \hline
\end{tabular}
\end{table*}

\begin{figure*}[hbt!]
\center{}
  \begin{subfigure}{.4\linewidth}
    \figuretitle{Average over time steps}
    \includegraphics[width=\linewidth]{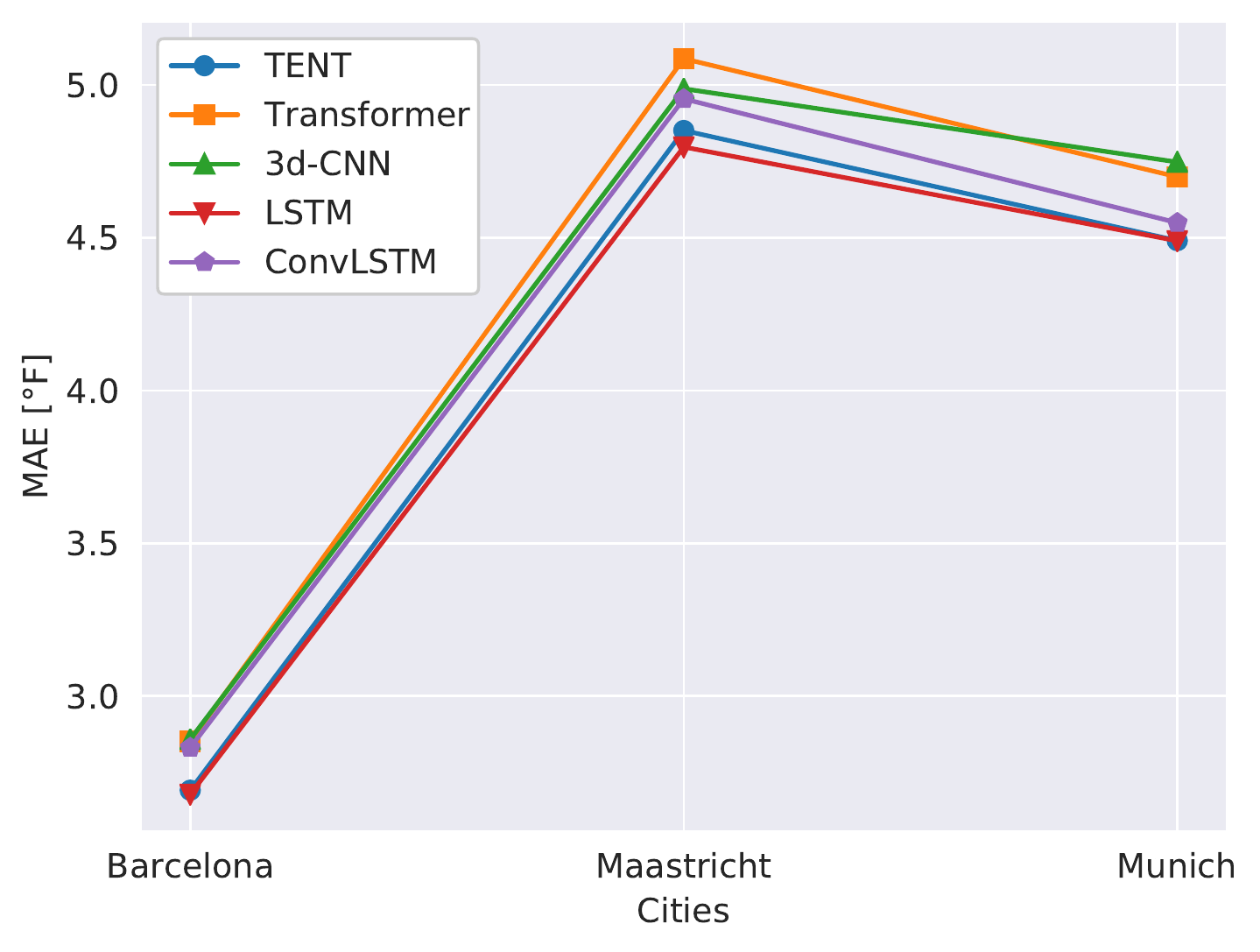}
    \caption{}
    \label{fig:mae_eu_aggregated_cities}
  \end{subfigure}\hspace{0.05\textwidth}
  \begin{subfigure}{.4\linewidth}
    \figuretitle{Average over cities}
    \includegraphics[width=\linewidth]{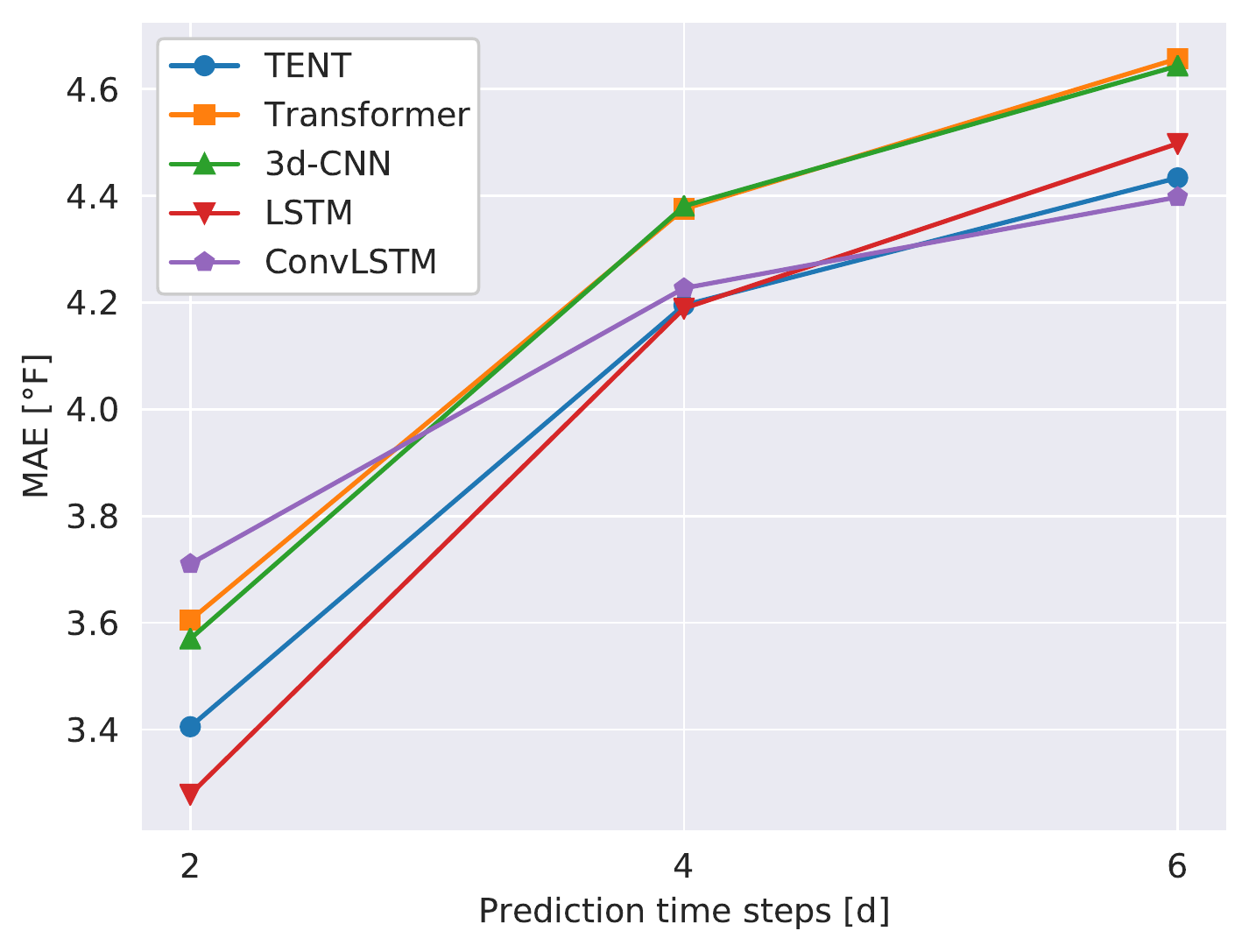}
    \caption{}
    \label{fig:mae_eu_aggregated_times}
  \end{subfigure}
  \caption{The obtained test MAE of the models for the \textbf{Europe} dataset averaged over cities (\ref{fig:mae_eu_aggregated_cities}) and prediction time steps (\ref{fig:mae_eu_aggregated_times}).}
  \label{fig:mae_eu_aggregated}
\end{figure*}

The longitude and latitude information of the cities are converted to Cartesian coordinates which are then used in normalized form as three additional features \cite{claessens2019efficient, hofmann2006physical} as follows:
\begin{equation}
\begin{cases}
      x = \cos(\phi) \cdot \cos(\lambda),\\
      y = \cos(\phi) \cdot \sin(\lambda),\\
      z = \sin(\phi).
\end{cases} 
\end{equation}
where $\phi$ and $\lambda$ are the latitude and longitude, respectively. Due to the time periodicity in the dataset, the day of the year and the hour of the day are added to each sample in the dataset and scaled along with the measurements in the dataset as in Eq. (\ref{eq:15}). The data from 2012-2016 is used as a training and validation set and the data from 2016-2017 forms the test set. 

\begin{equation} \label{eq:15}
x_{\textrm{scaled}} = \frac{x-\min(x)}{\max(x)-\min(x)}.
\end{equation}

We cast the input data to a tensor with the shape $T \times C \times F$, where the first, second and third dimensions represent the time sequence, cities and the features of the cities respectively. Predictions are made for 4, 8, 12 and 16 hours into the future (prediction time step). In our conducted experiments the lag of $16$ hours is used as it was empirically found to be optimal in terms of the mean squared error on the validation set. For this dataset, the target cities are Vancouver, Dallas and New York and the target feature is temperature. The obtained test MAEs and MSEs are tabulated in Table \ref{tab:usa_results}. Fig. \ref{fig:mae_usa_aggregated} (a) and (b) show the averaged MAE over cities and time steps, respectively. A subset of the obtained test prediction and real measurement for Vancouver city for 4, 8, 12 and 16 hours ahead are also depicted in Fig. \ref{fig:PredRealEUA}. As can be seen, the prediction accuracy decreases as the number of hours ahead increases. The obtained test MAE and MSE results exhibit consistent pattern, with the TENT model outperforming other tested models for all tested time step ahead predictions. Our model also obtained the lowest average MAE and MSE over both cities and prediction time steps.

\subsection{Europe Dataset}
This dataset contains daily measurements of the weather attributes of 18 cities across Europe from May 2005 to April 2020. The time periodicity is added in the form of the day of the year and normalized as other features. The data from 2005-2017 is used for training and validation and the data from 2017-2020 is used as the test set. 

\begin{figure*}[hbt!]
\center{}
  \begin{subfigure}{0.25\linewidth}
    \figuretitle{4 hours into the future}
    \includegraphics[width=\linewidth]{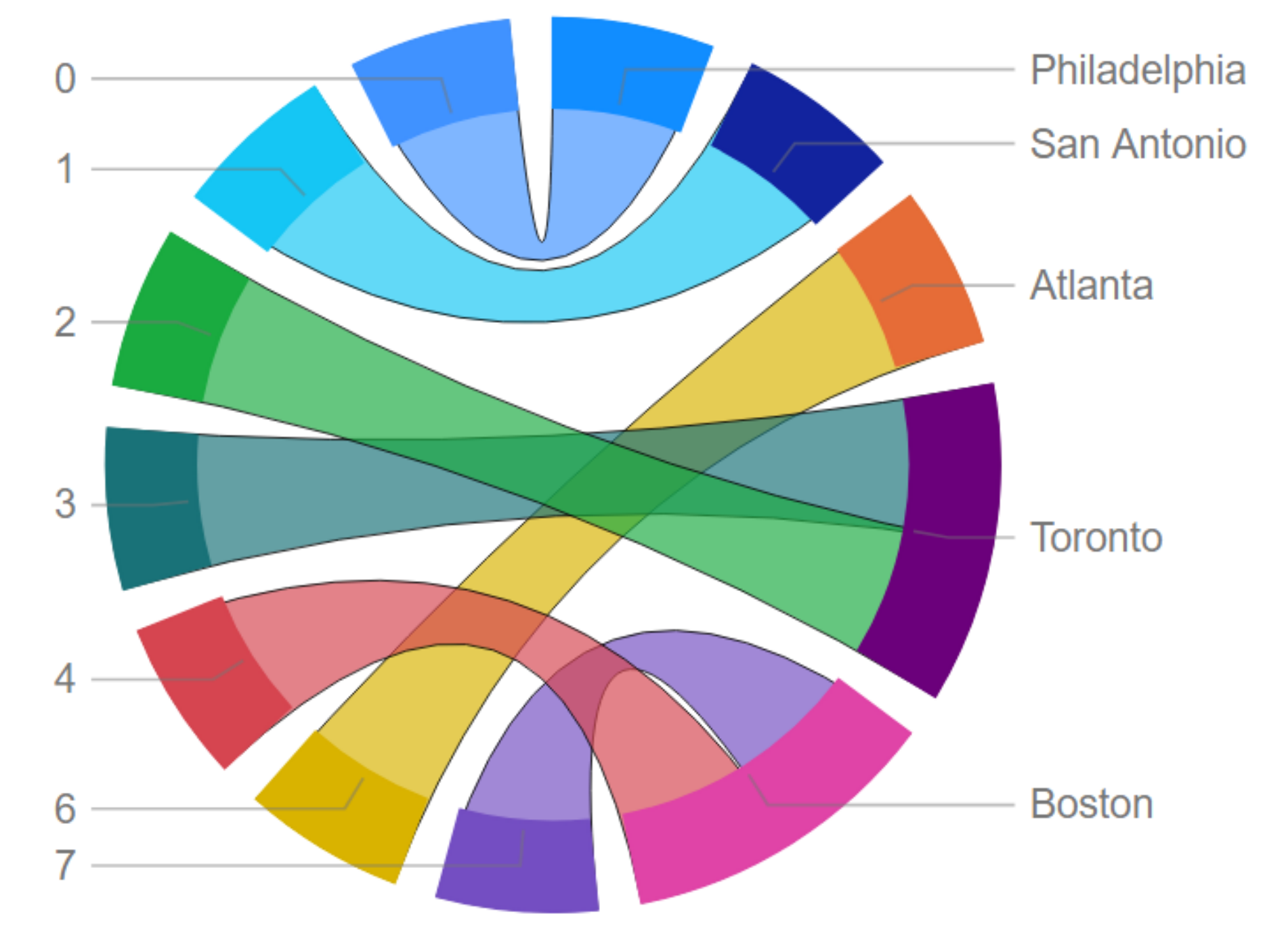}
    \caption{}\vspace{1em}
    \label{fig:DallasCircular_lag4}
  \end{subfigure}\hfill
  \begin{subfigure}{0.25\linewidth}
    \figuretitle{8 hours into the future}
    \includegraphics[width=\linewidth]{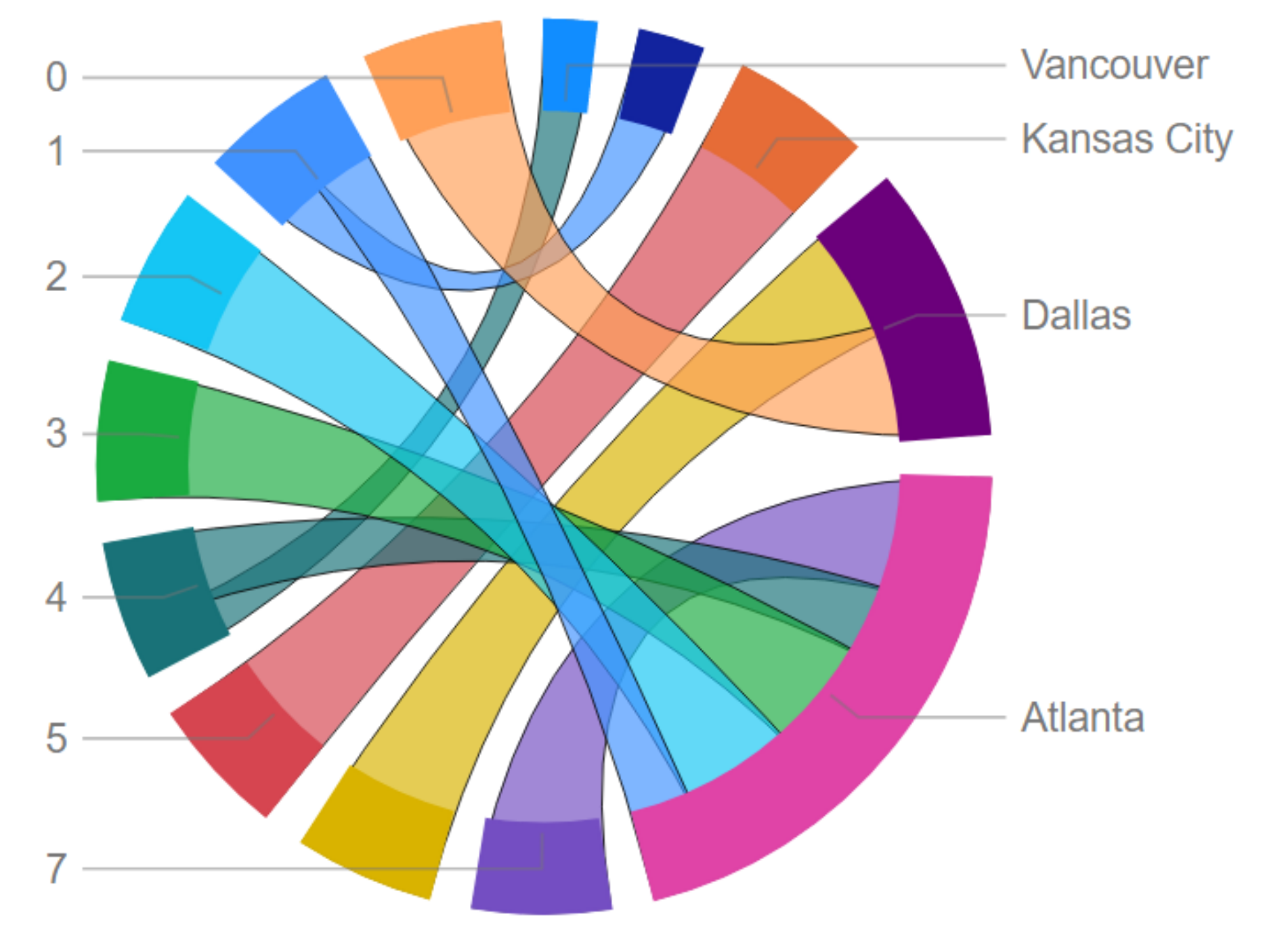}
    \caption{}\vspace{1em}
    \label{fig:DallasCircular_lag8}
  \end{subfigure}\hfill
  \begin{subfigure}{0.25\linewidth}
    \figuretitle{12 hours into the future}
    \includegraphics[width=\linewidth]{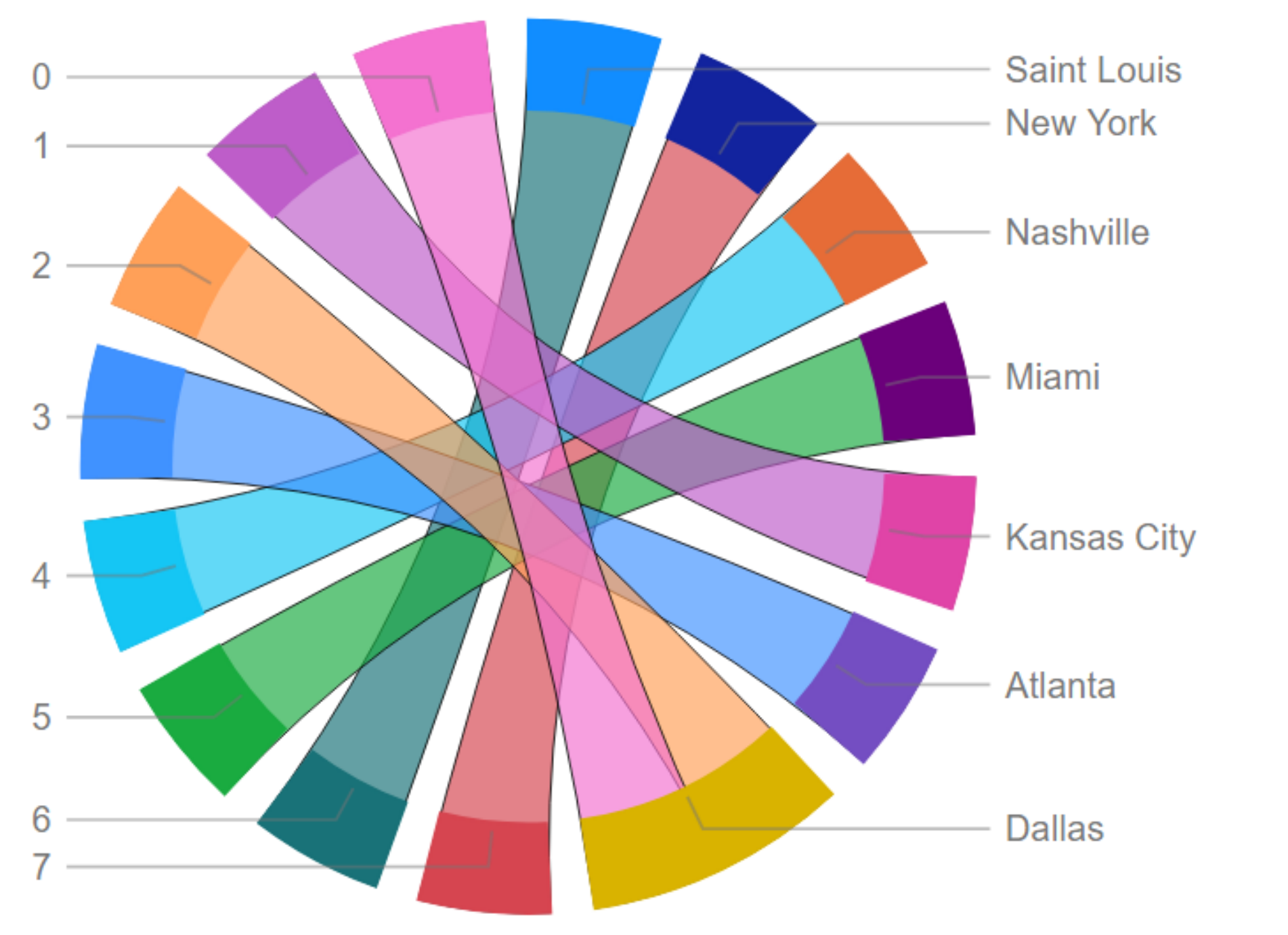}
    \caption{}\vspace{1em}
    \label{fig:DallasCircular_lag12}
  \end{subfigure}\hfill
  \begin{subfigure}{0.25\linewidth}
    \figuretitle{16 hours into the future}
    \includegraphics[width=\linewidth]{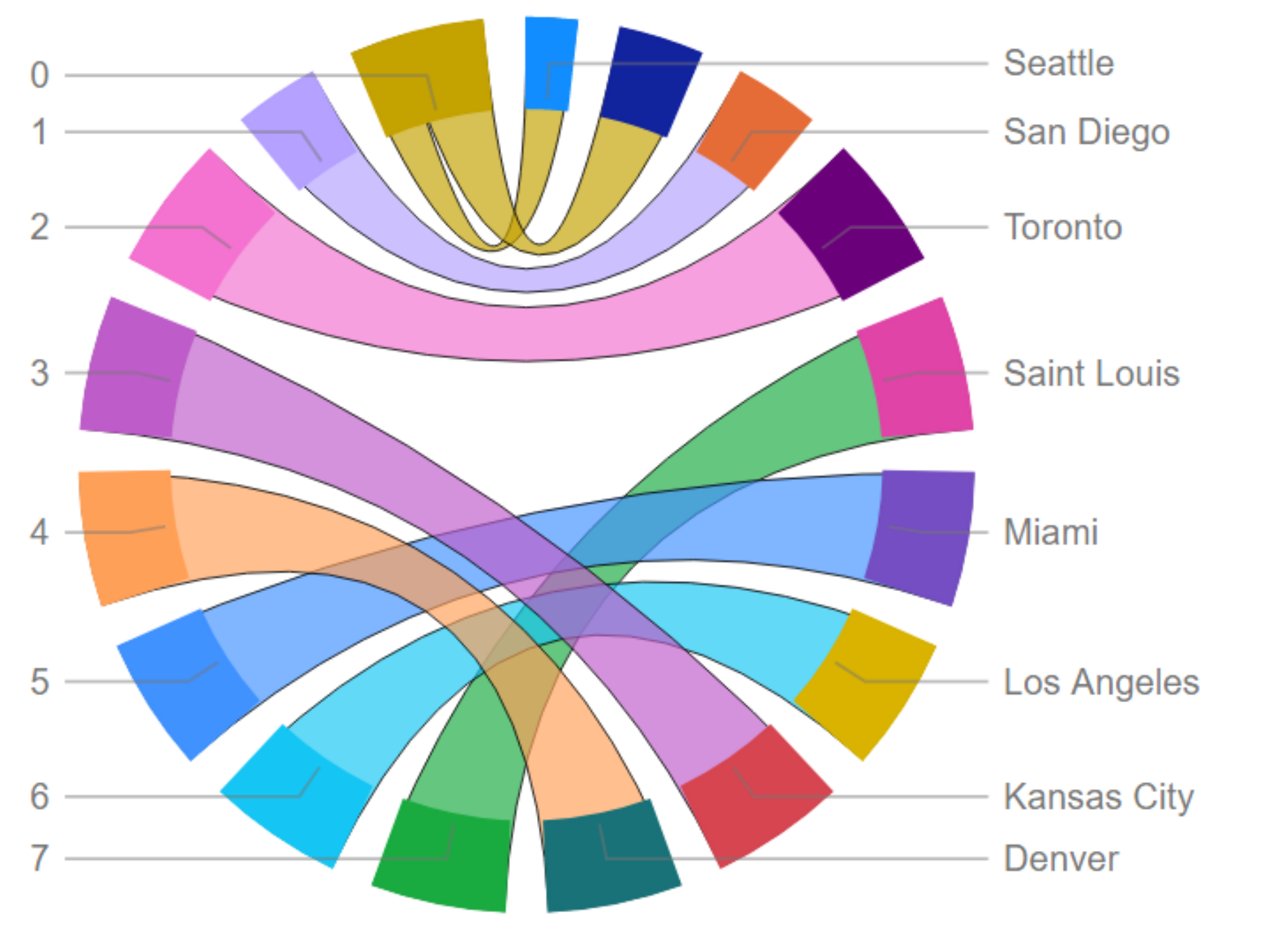}
    \caption{}\vspace{1em}
    \label{fig:DallasCircular_lag16}
  \end{subfigure}\hfill
    \begin{subfigure}{0.24\linewidth}
    \figuretitle{4 hours into the future}
    \includegraphics[width=\linewidth]{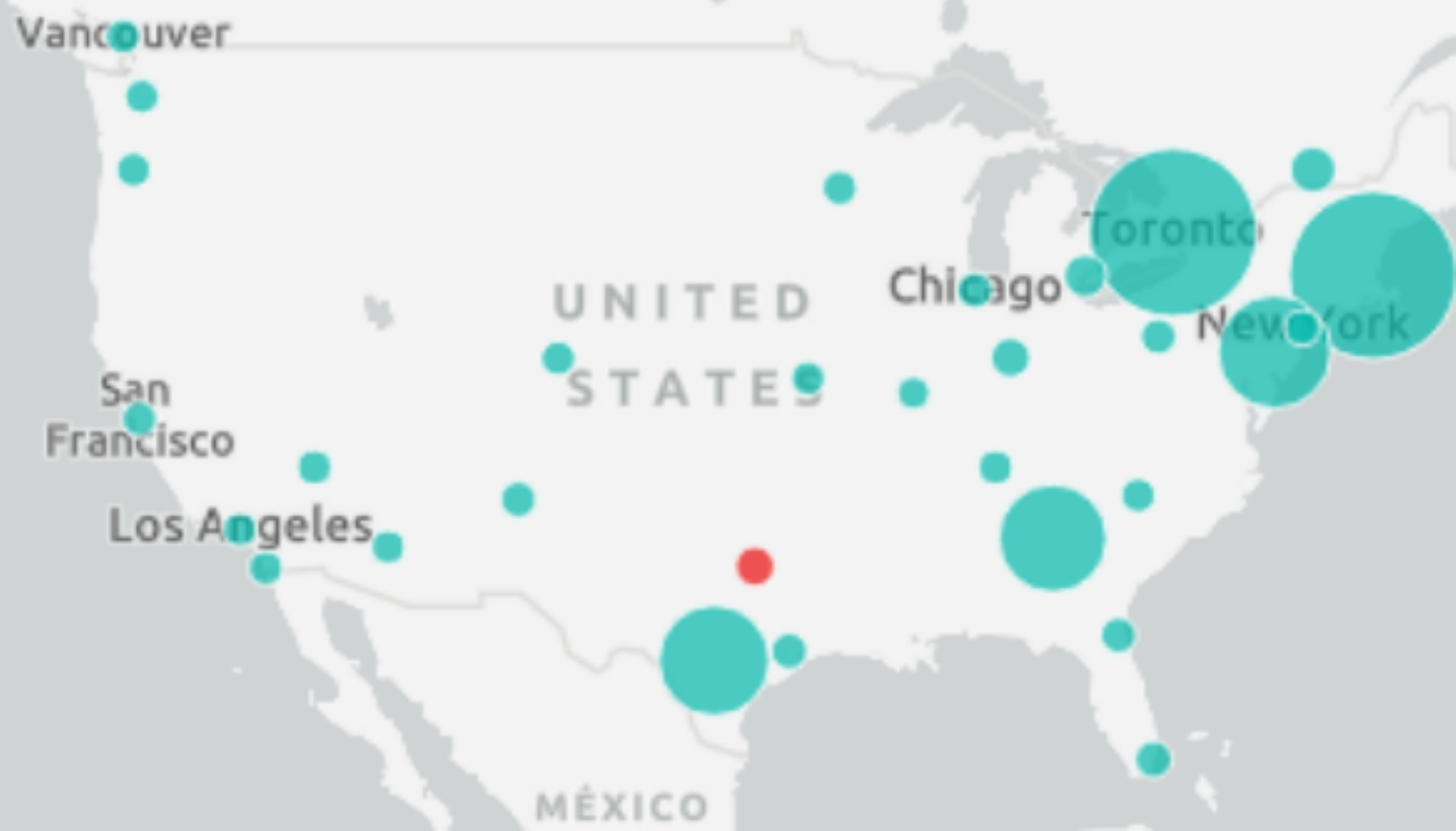}
    \caption{}
    \label{fig:DallasMaps_lag4}
  \end{subfigure}\hspace{0.001\textwidth}
  \begin{subfigure}{0.24\linewidth}
    \figuretitle{8 hours into the future}
    \includegraphics[width=\linewidth]{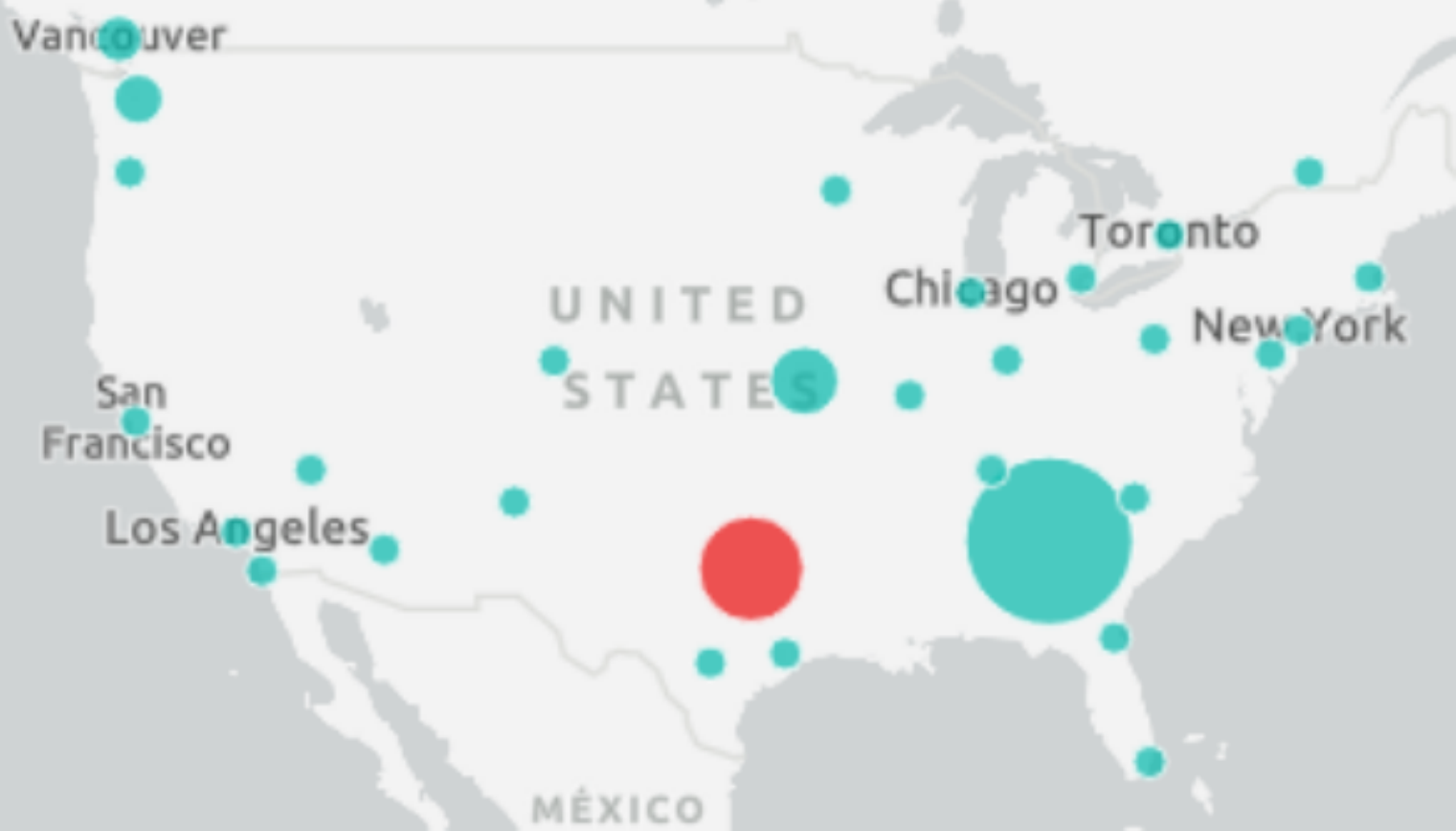}
    \caption{}
    \label{fig:DallasMaps_lag8}
  \end{subfigure}\hspace{0.001\textwidth}
  \begin{subfigure}{0.24\linewidth}
    \figuretitle{12 hours into the future}
    \includegraphics[width=\linewidth]{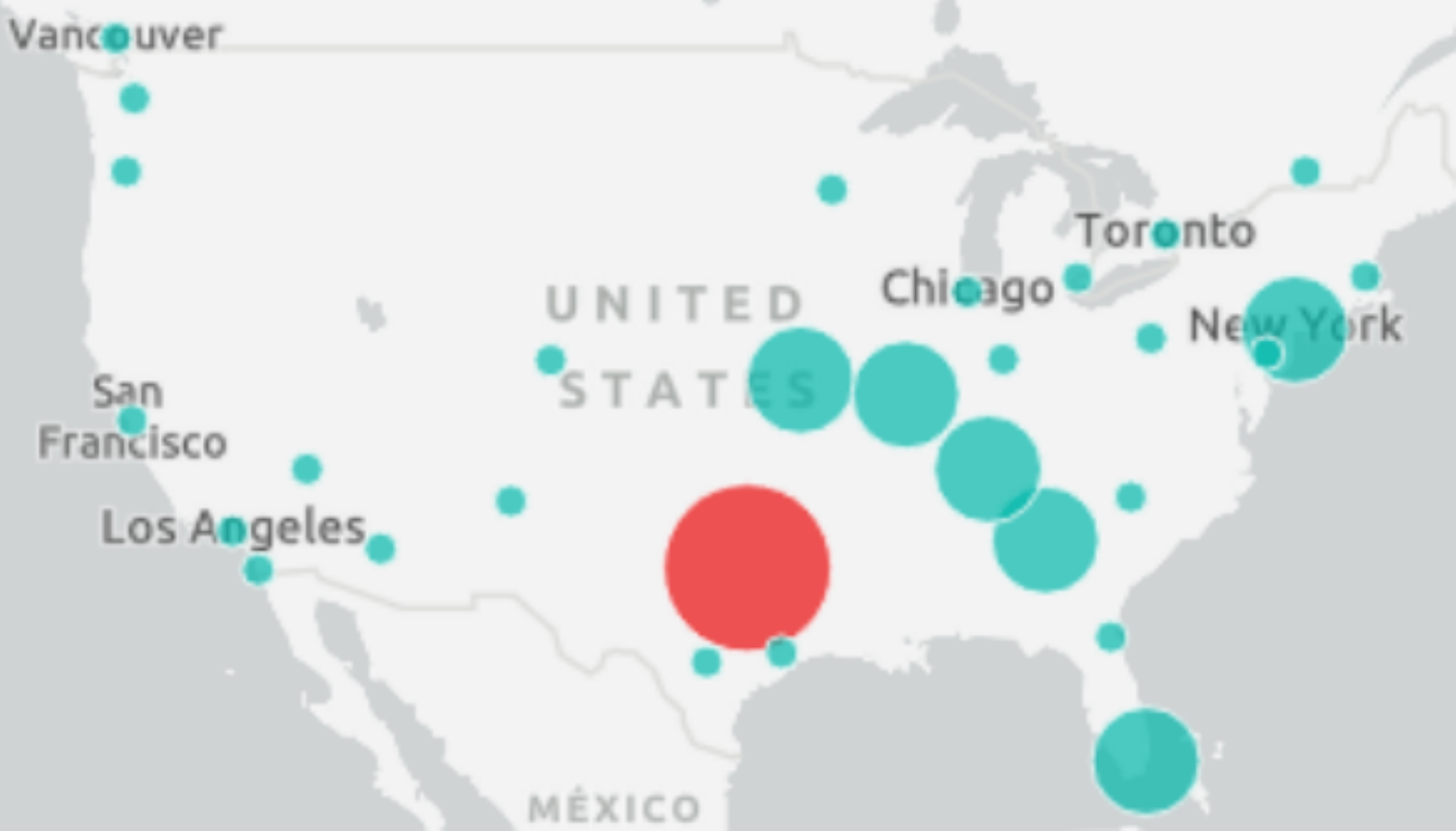}
    \caption{}
    \label{fig:DallasMaps_lag12}
  \end{subfigure}\hspace{0.001\textwidth}
  \begin{subfigure}{0.24\linewidth}
    \figuretitle{16 hours into the future}
    \includegraphics[width=\linewidth]{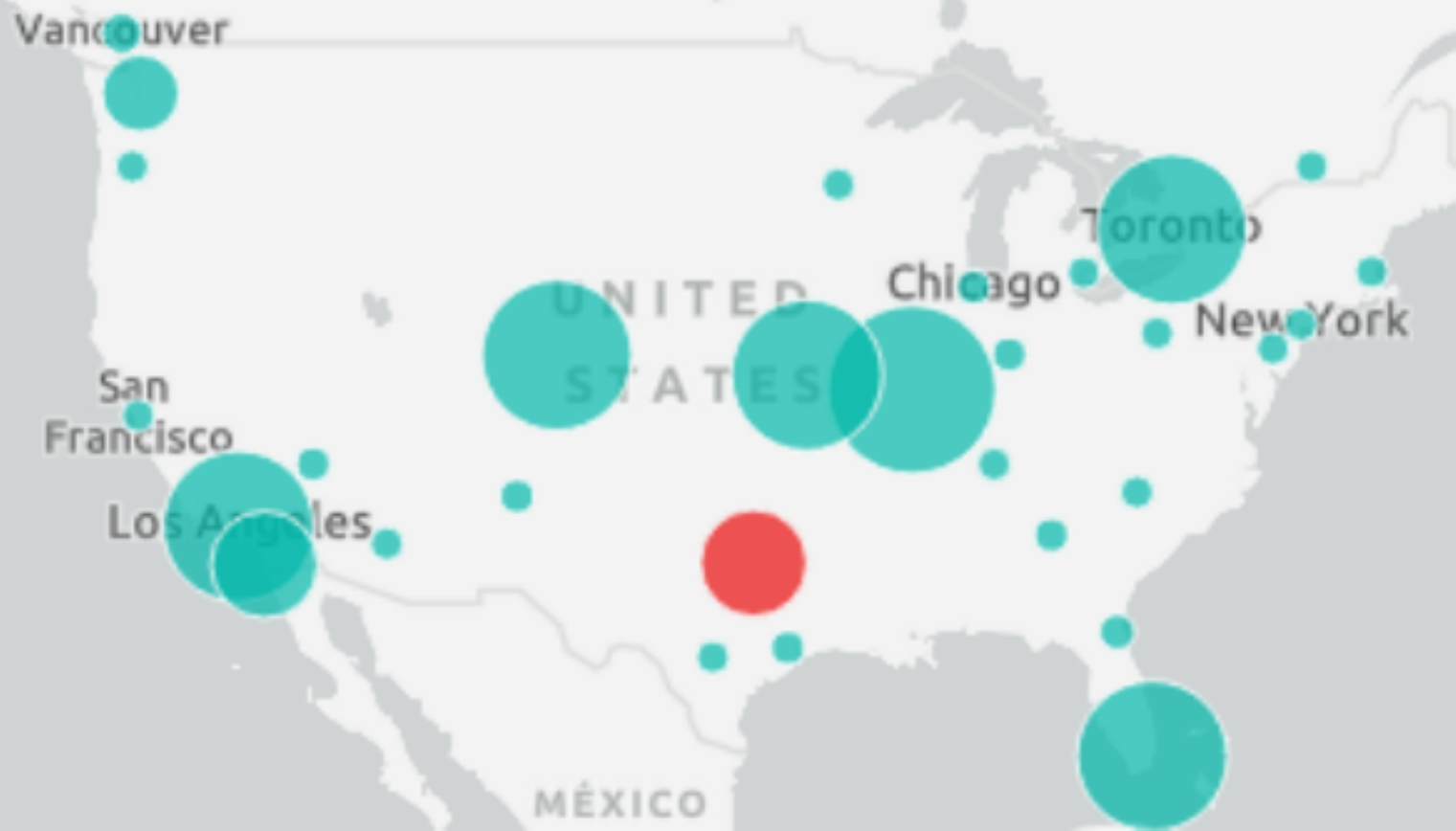}
    \caption{}
    \label{fig:DallasMaps_lag16}
  \end{subfigure}
  \caption{Attention visualization for \textbf{Dallas} in USA-Canada dataset. The circular graphs shows which city each of the most important heads attends to. The thickness of the line represents the amount of attention each of the heads is paying to the cities. The size of the circles indicates the importance of
each city in the temperature prediction for the target city. The target city is marked as a red circle and its size corresponds to the importance of the attention to itself.}
  \label{fig:Dallas}
\end{figure*}

In this dataset, we perform experiments for 2, 4 and 6 days ahead prediction. The optimal lag parameter used to construct the regressor is empirically found and is set to 8 days. The target cities are Barcelona, Maastricht and Munich and the target feature is the average temperature. The obtained MAEs and MSEs are tabulated in Table \ref{tab:eu_results}. Fig. \ref{fig:mae_eu_aggregated} (a) and (b) shows the MAE of the models averaged over cities and prediction time steps respectively. 
In the experiments, in general, the proposed TENT model is the second best model and only outperforms the other tested models when predicting the temperature of Maastricht city for 4 and 6 days ahead.
In this dataset, LSTM has the highest performance among other models, i.e. it achieved the best MAE on 5 prediction tasks (a particular city and a particular prediction time) and the best MSE in 4 of them. 
It has been previously reported by \cite{guo2019star, ezen2020comparison}, that Transformer might not reach its performance capacity when dealing with the small data size. Therefore, we think that the limited data in the EU dataset potentially prevents TENT to achieve better results compared to the LSTM model.

\begin{figure*}[hbt!]
\center{}
  \begin{subfigure}{0.25\linewidth}
    \figuretitle{4 hours into the future}
    \includegraphics[width=\linewidth]{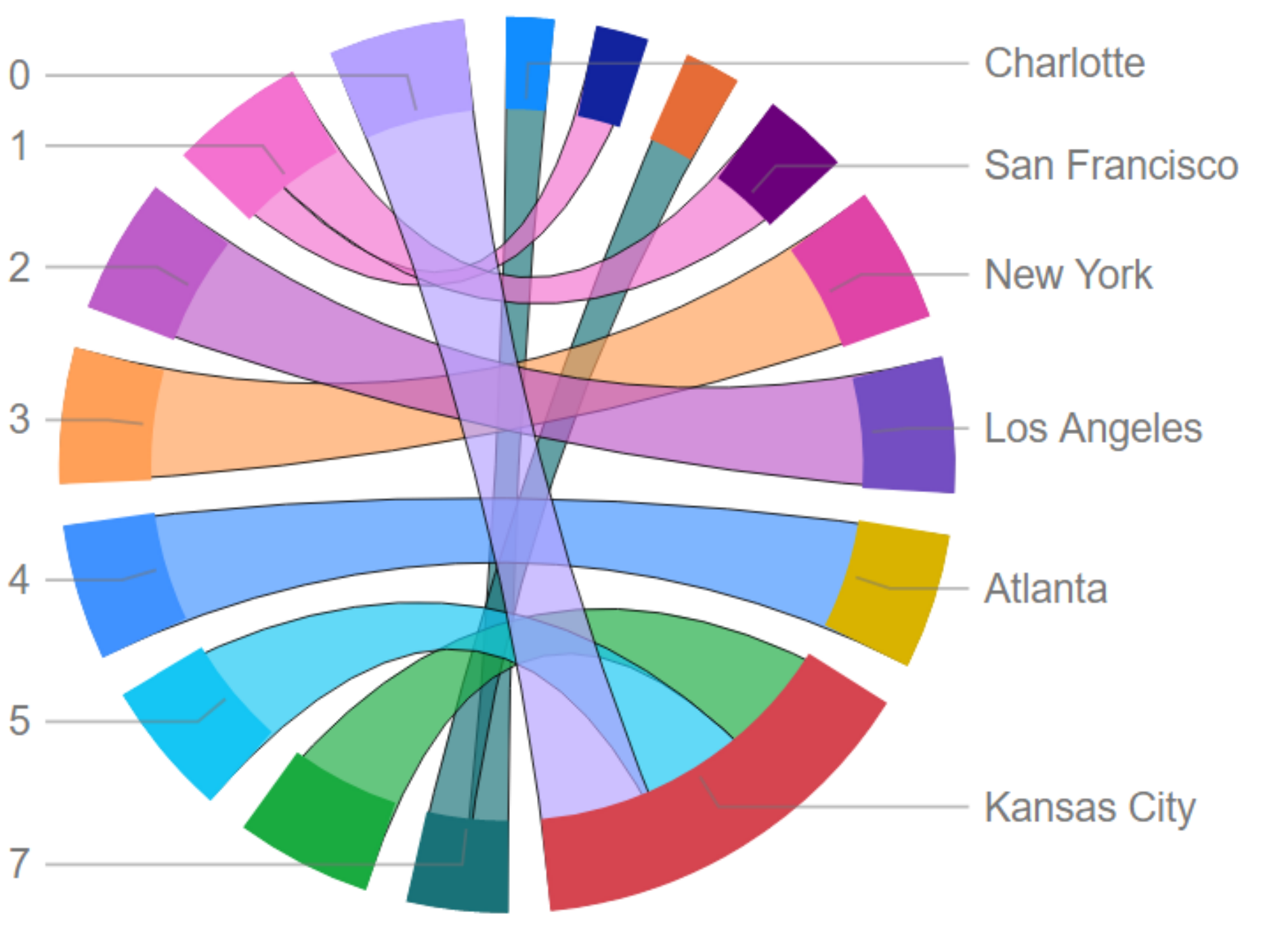}
    \caption{}\vspace{1em}
    \label{fig:VancouverCircular_lag4}
  \end{subfigure}\hfill
  \begin{subfigure}{0.25\linewidth}
    \figuretitle{8 hours into the future}
    \includegraphics[width=\linewidth]{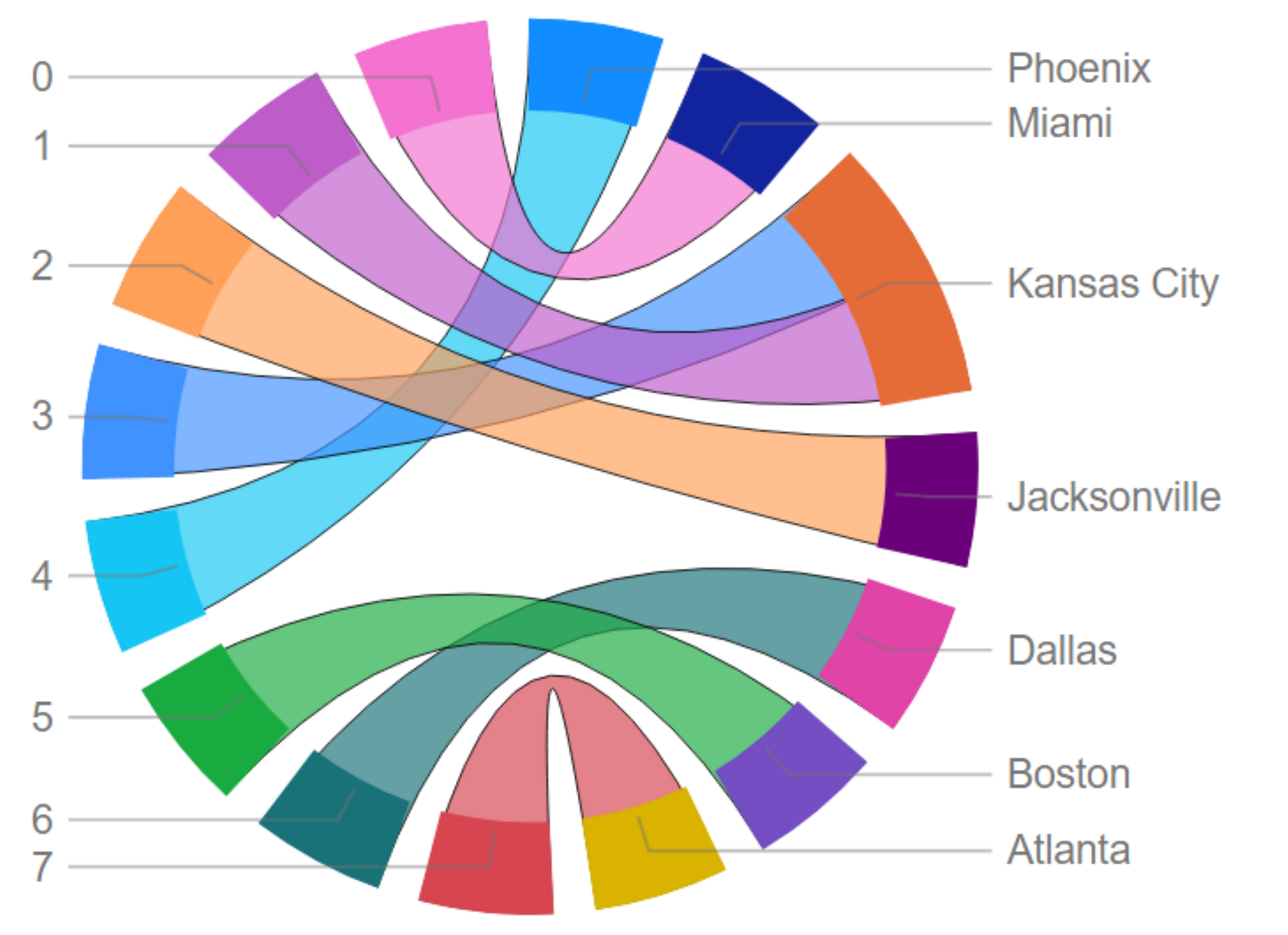}
    \caption{}\vspace{1em}
    \label{fig:VancouverCircular_lag8}
  \end{subfigure}\hfill
  \begin{subfigure}{0.25\linewidth}
    \figuretitle{12 hours into the future}
    \includegraphics[width=\linewidth]{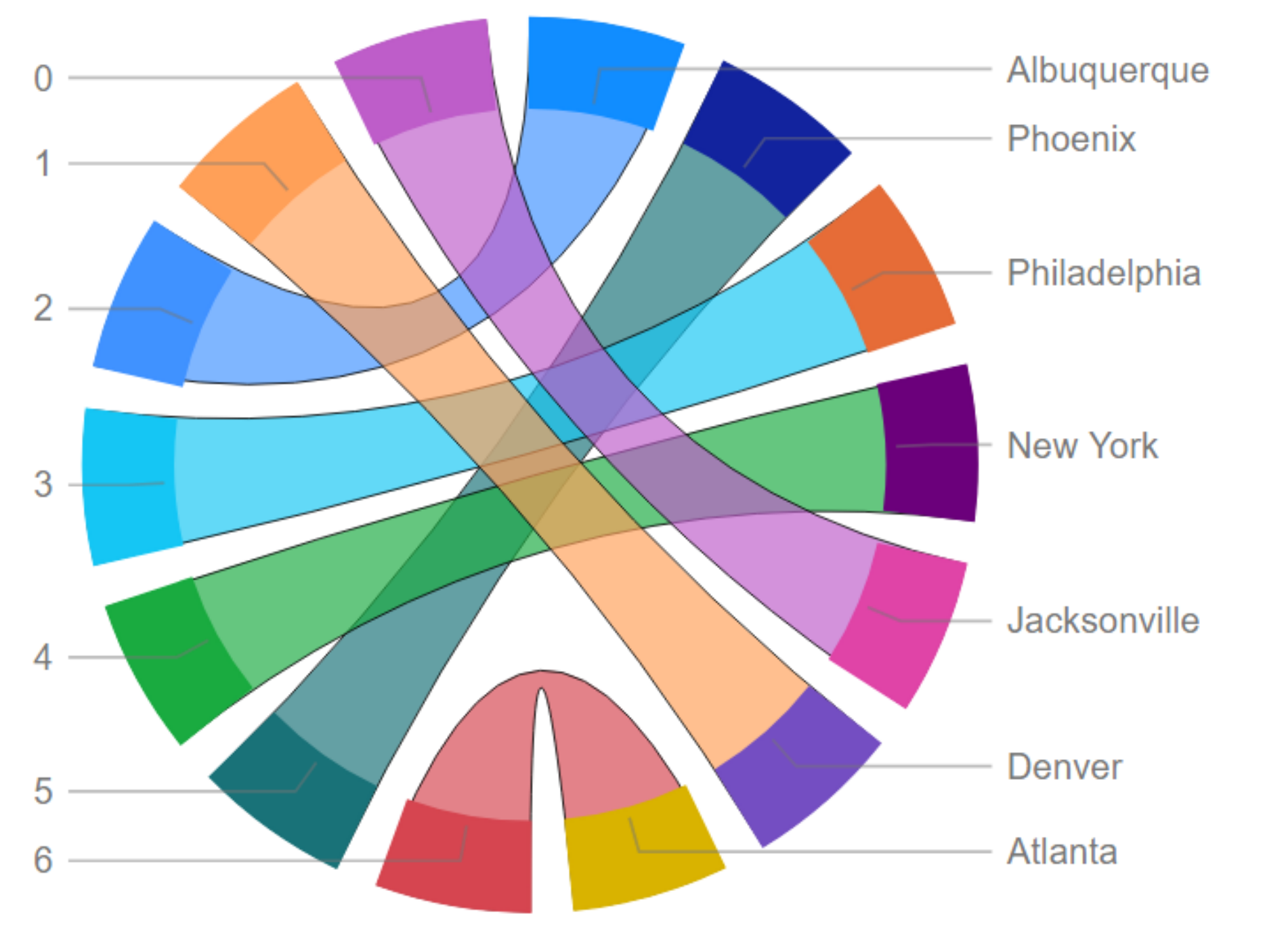}
    \caption{}\vspace{1em}
    \label{fig:VancouverCircular_lag12}
  \end{subfigure}\hfill
  \begin{subfigure}{0.25\linewidth}
    \figuretitle{16 hours into the future}
    \includegraphics[width=\linewidth]{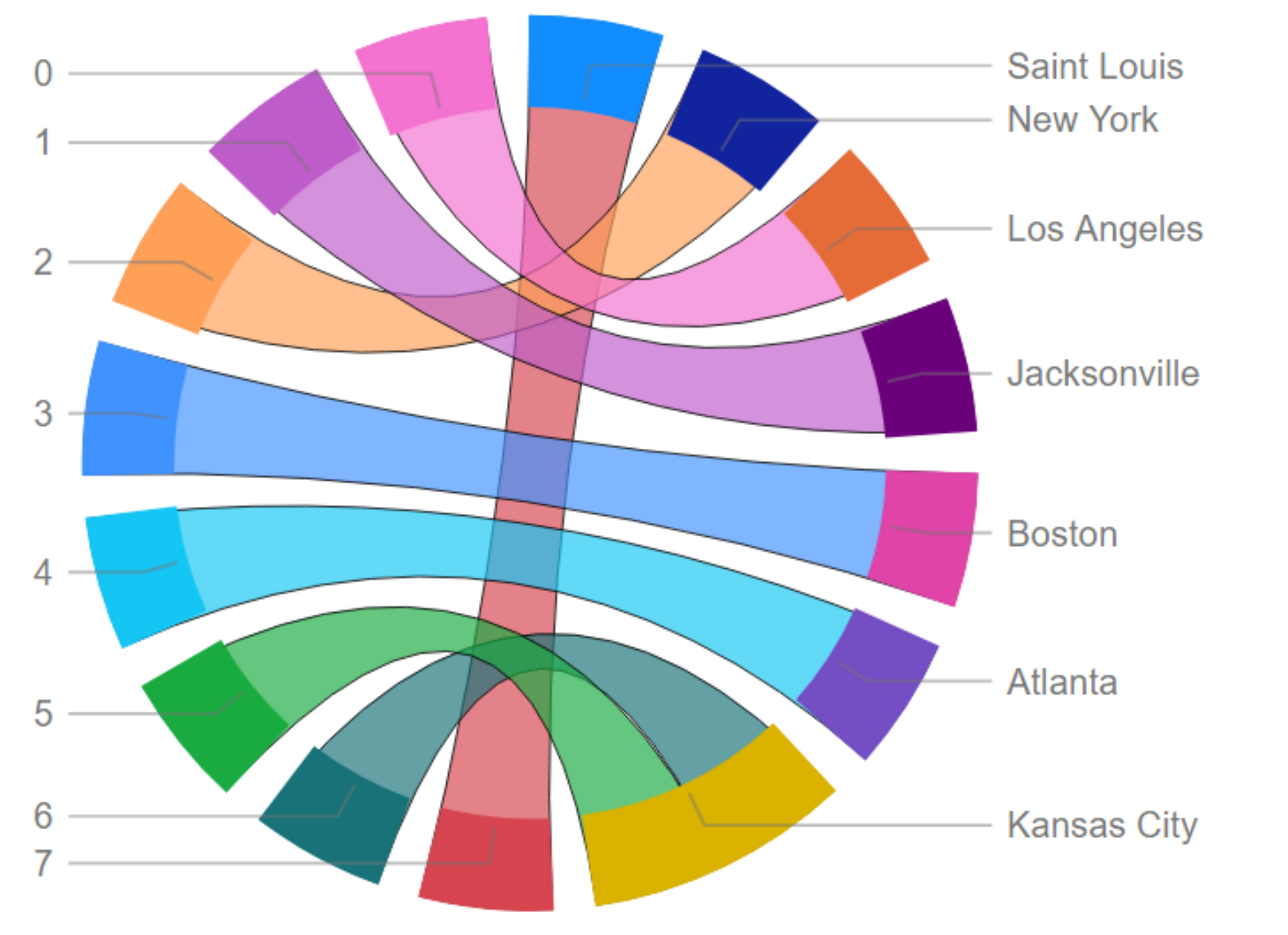}
    \caption{}\vspace{1em}
    \label{fig:VancouverCircular_lag16}
  \end{subfigure}\hfill
   \begin{subfigure}{0.24\linewidth}
    \figuretitle{4 hours into the future}
    \includegraphics[width=\linewidth]{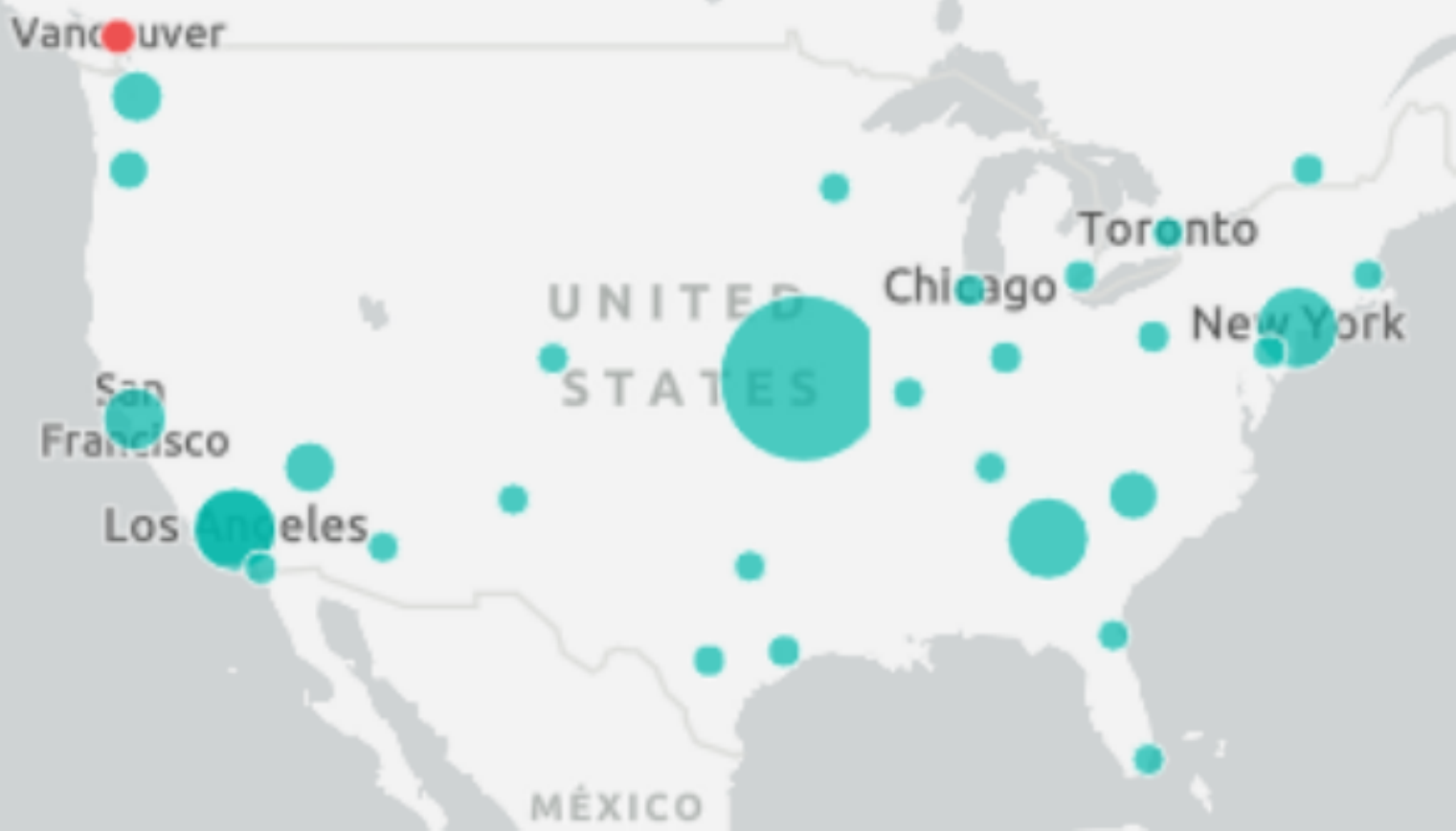}
    \caption{}
    \label{fig:VancouverMaps_lag4}
  \end{subfigure}\hspace{0.001\textwidth}
  \begin{subfigure}{0.24\linewidth}
    \figuretitle{8 hours into the future}
    \includegraphics[width=\linewidth]{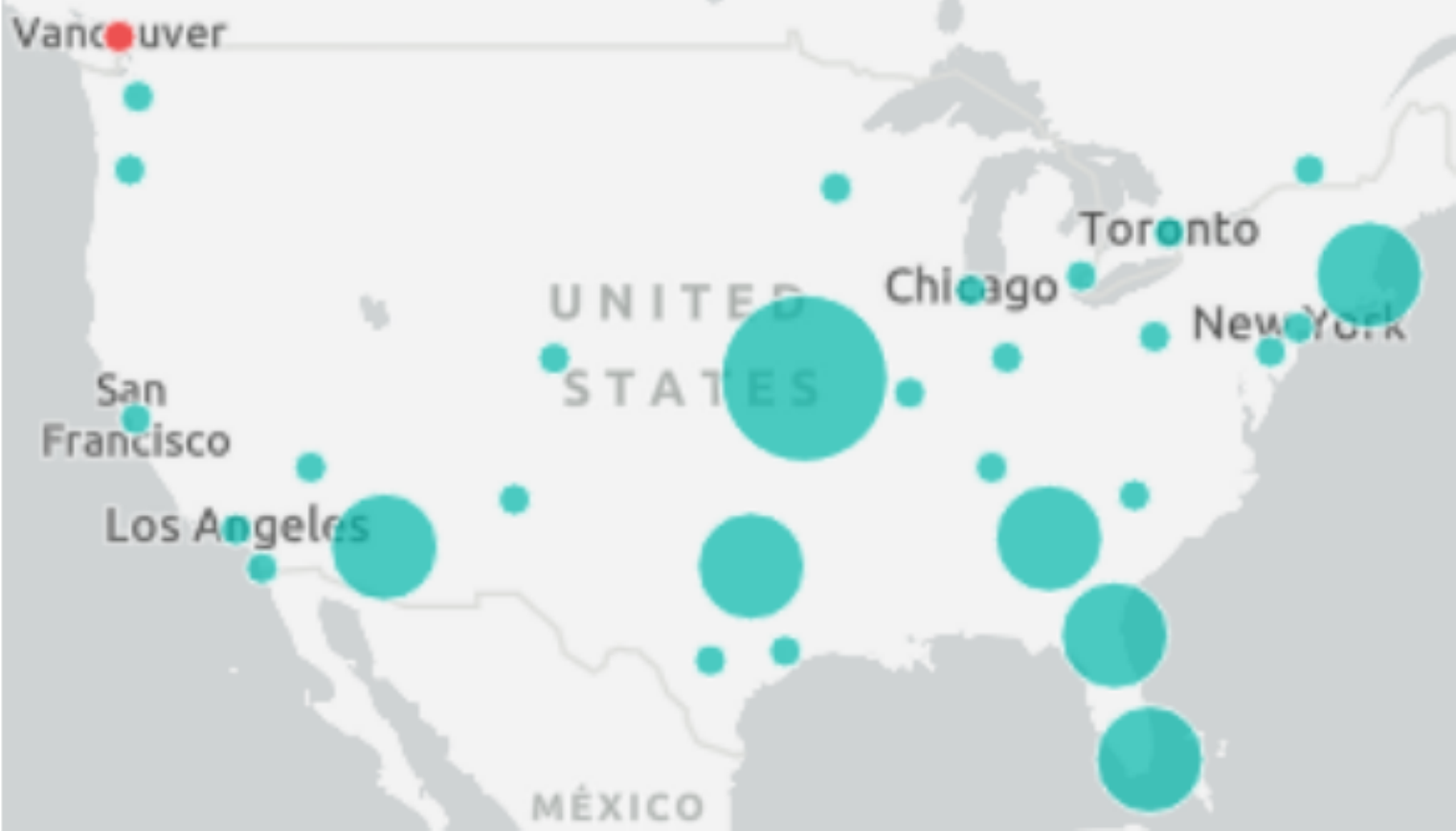}
    \caption{}
    \label{fig:VancouverMaps_lag8}
  \end{subfigure}\hspace{0.001\textwidth}
  \begin{subfigure}{0.24\linewidth}
    \figuretitle{12 hours into the future}
    \includegraphics[width=\linewidth]{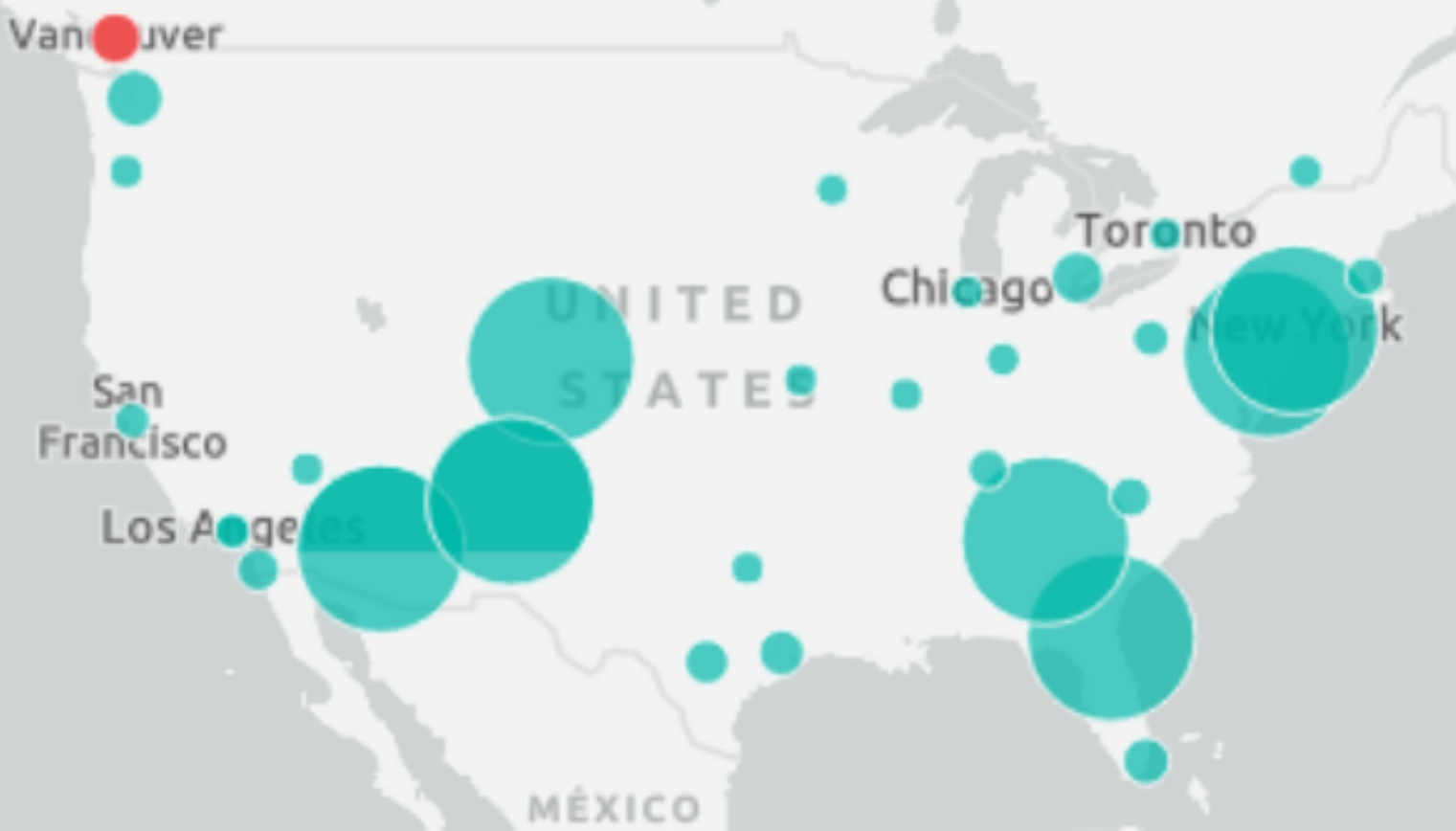}
    \caption{}
    \label{fig:VancouverMaps_lag12}
  \end{subfigure}\hspace{0.001\textwidth}
  \begin{subfigure}{0.24\linewidth}
    \figuretitle{16 hours into the future}
    \includegraphics[width=\linewidth]{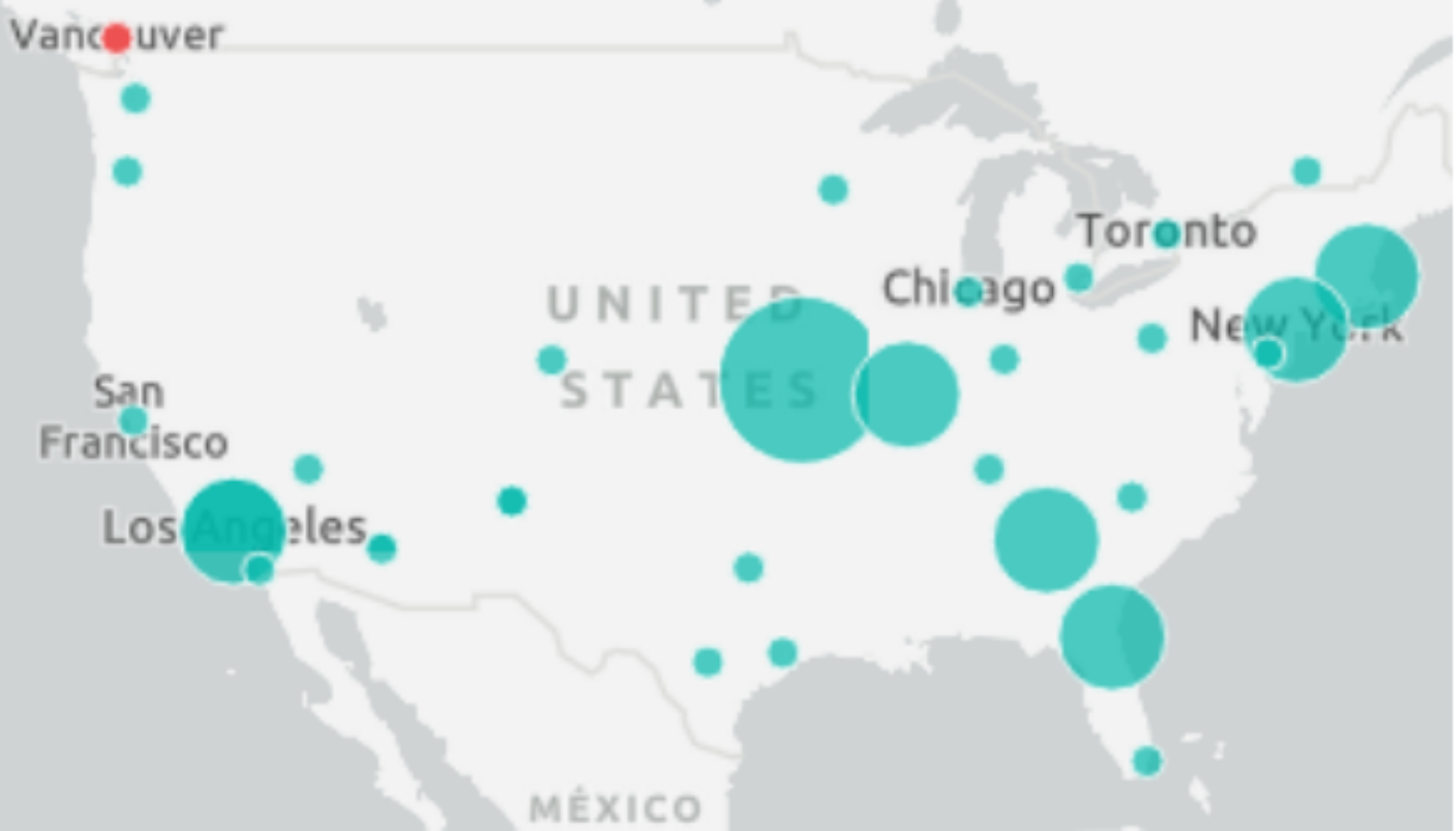}
    \caption{}
    \label{fig:VancouverMaps_lag16}
  \end{subfigure}
  \caption{Attention visualization for \textbf{Vancouver} in USA-Canada dataset. The circular graphs shows which city each of the most important heads attends to. The thickness of the line represents the amount of attention each of the heads is paying to the cities. The size of the circles indicates the importance of
each city in the temperature prediction for the target city. The target city is marked as a red circle and its size corresponds to the importance of the attention to itself.}
  \label{fig:Vancouver}
\end{figure*}

\begin{figure*}[hbt!]
\center{}
  \begin{subfigure}{0.25\linewidth}
    \figuretitle{4 hours into the future}
    \includegraphics[width=\linewidth]{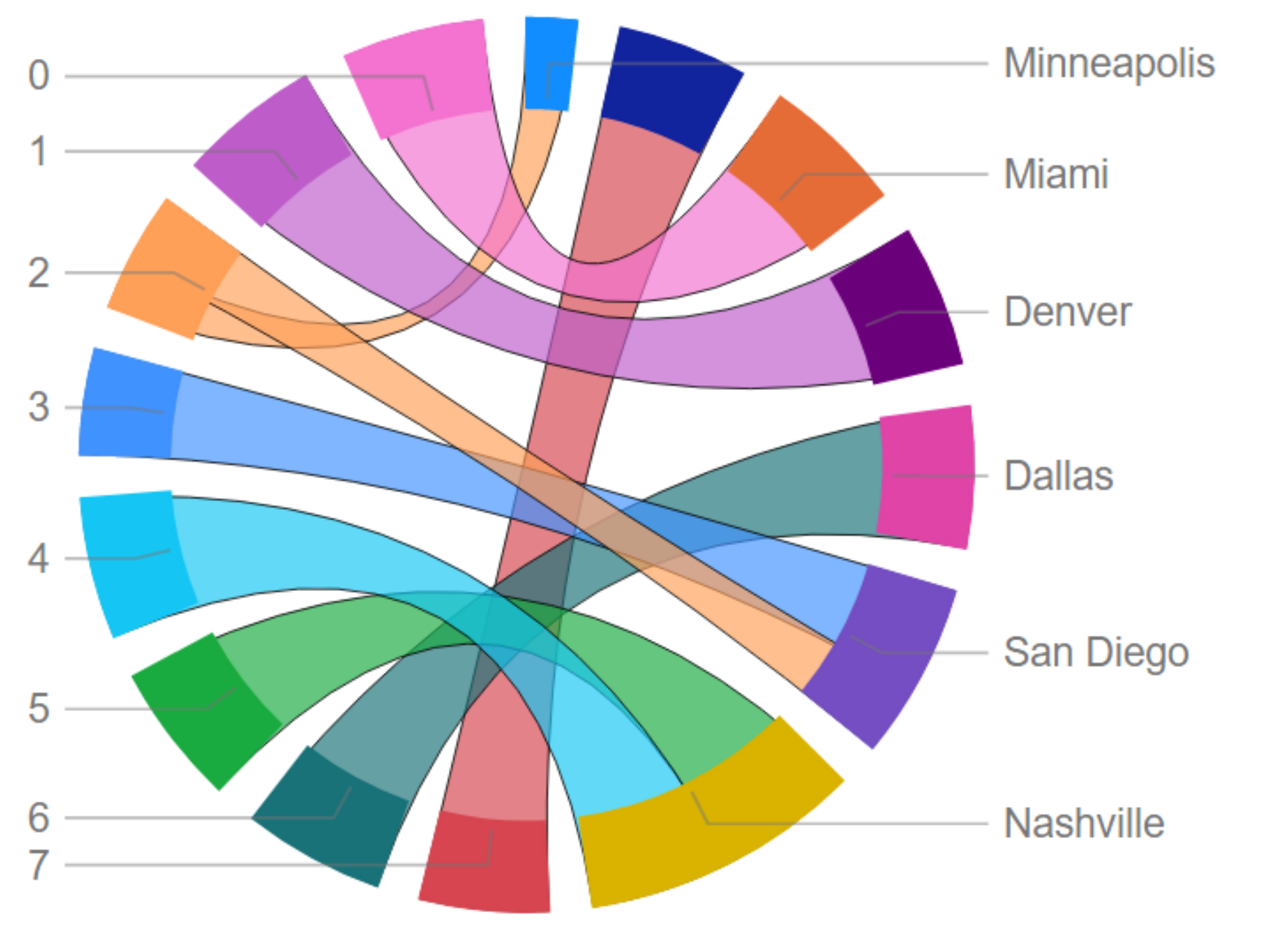}
    \caption{}\vspace{1em}
    \label{fig:NewYorkCircular_lag4}
  \end{subfigure}\hfill
  \begin{subfigure}{0.25\linewidth}
    \figuretitle{8 hours into the future}
    \includegraphics[width=\linewidth]{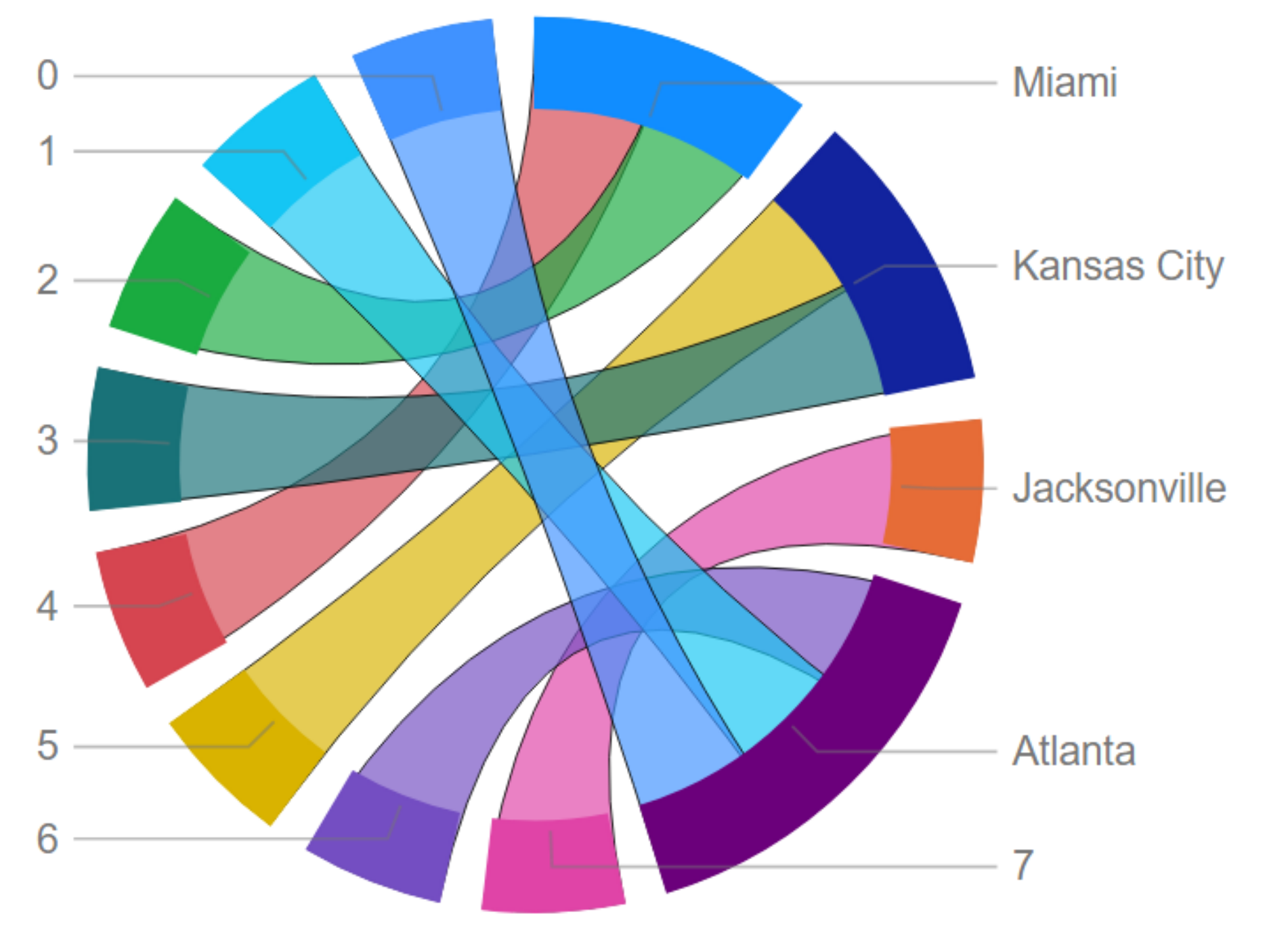}
    \caption{}\vspace{1em}
    \label{fig:NewYorkCircular_lag8}
  \end{subfigure}\hfill
  \begin{subfigure}{0.25\linewidth}
    \figuretitle{12 hours into the future}
    \includegraphics[width=\linewidth]{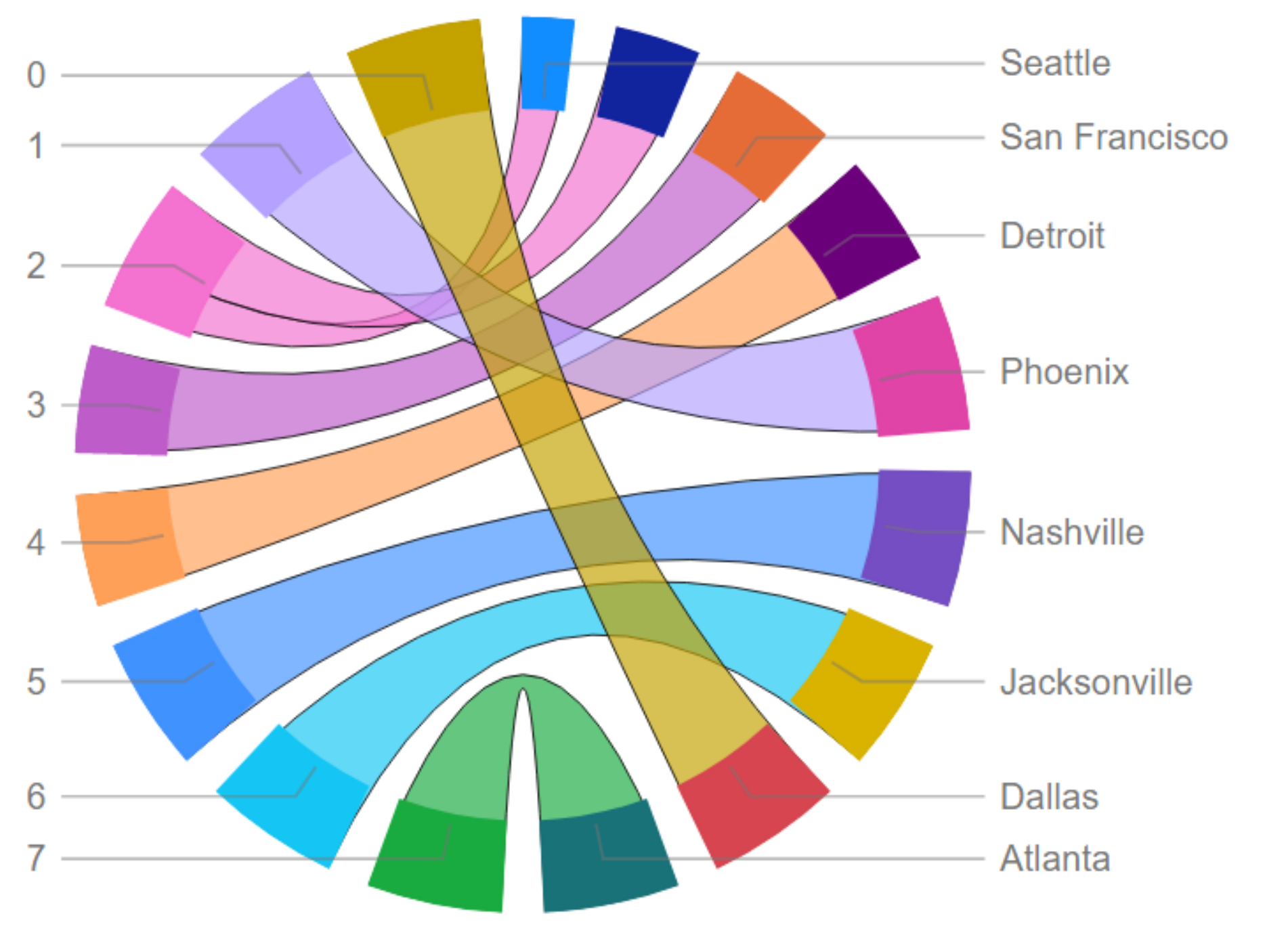}
    \caption{}\vspace{1em}
    \label{fig:NewYorkCircular_lag12}
  \end{subfigure}\hfill
  \begin{subfigure}{0.25\linewidth}
    \figuretitle{16 hours into the future}
    \includegraphics[width=\linewidth]{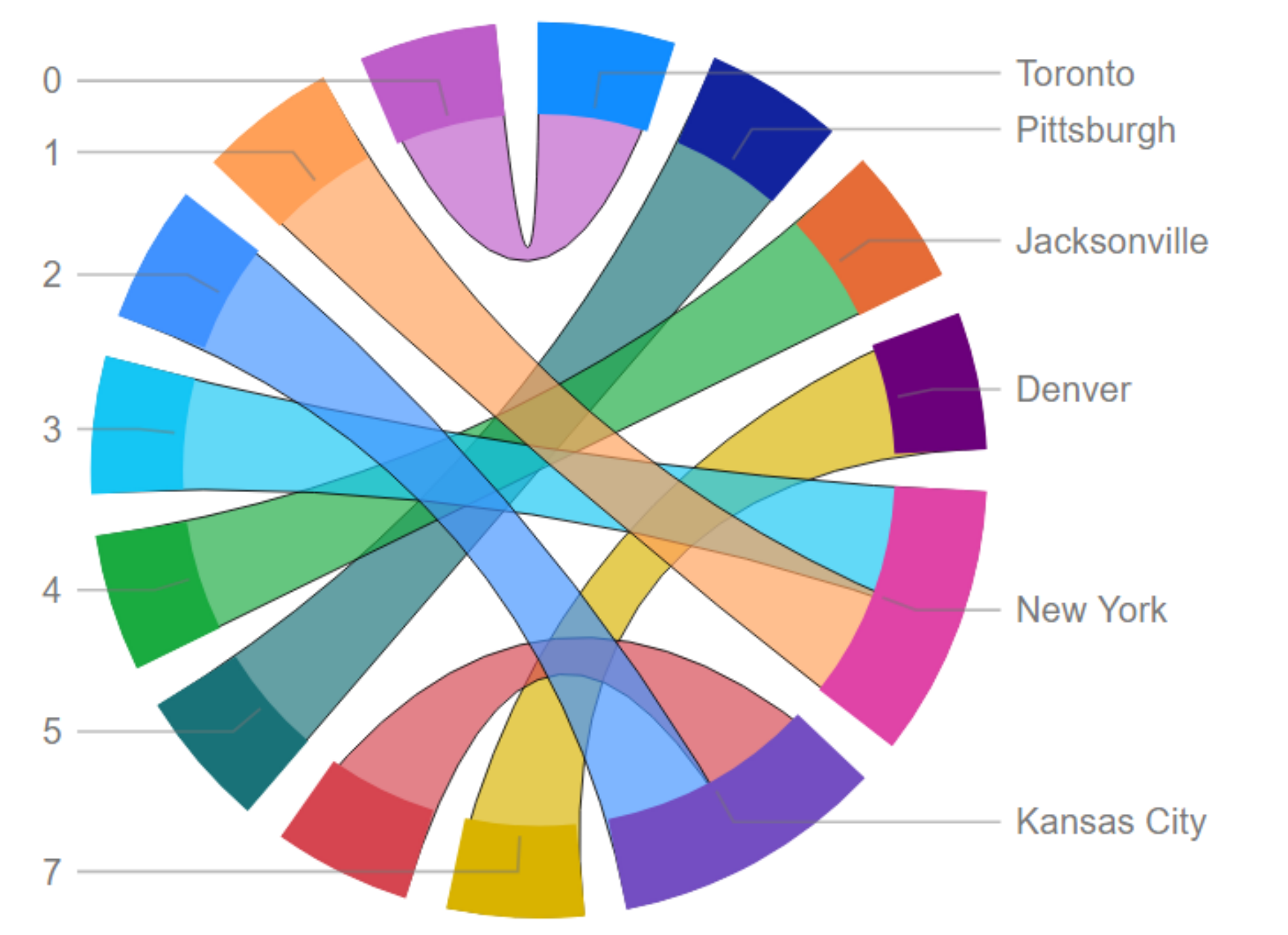}
    \caption{}\vspace{1em}
    \label{fig:NewYorkCircular_lag16}
  \end{subfigure}\hfill
    \begin{subfigure}{0.24\linewidth}
    \figuretitle{4 hours into the future}
    \includegraphics[width=\linewidth]{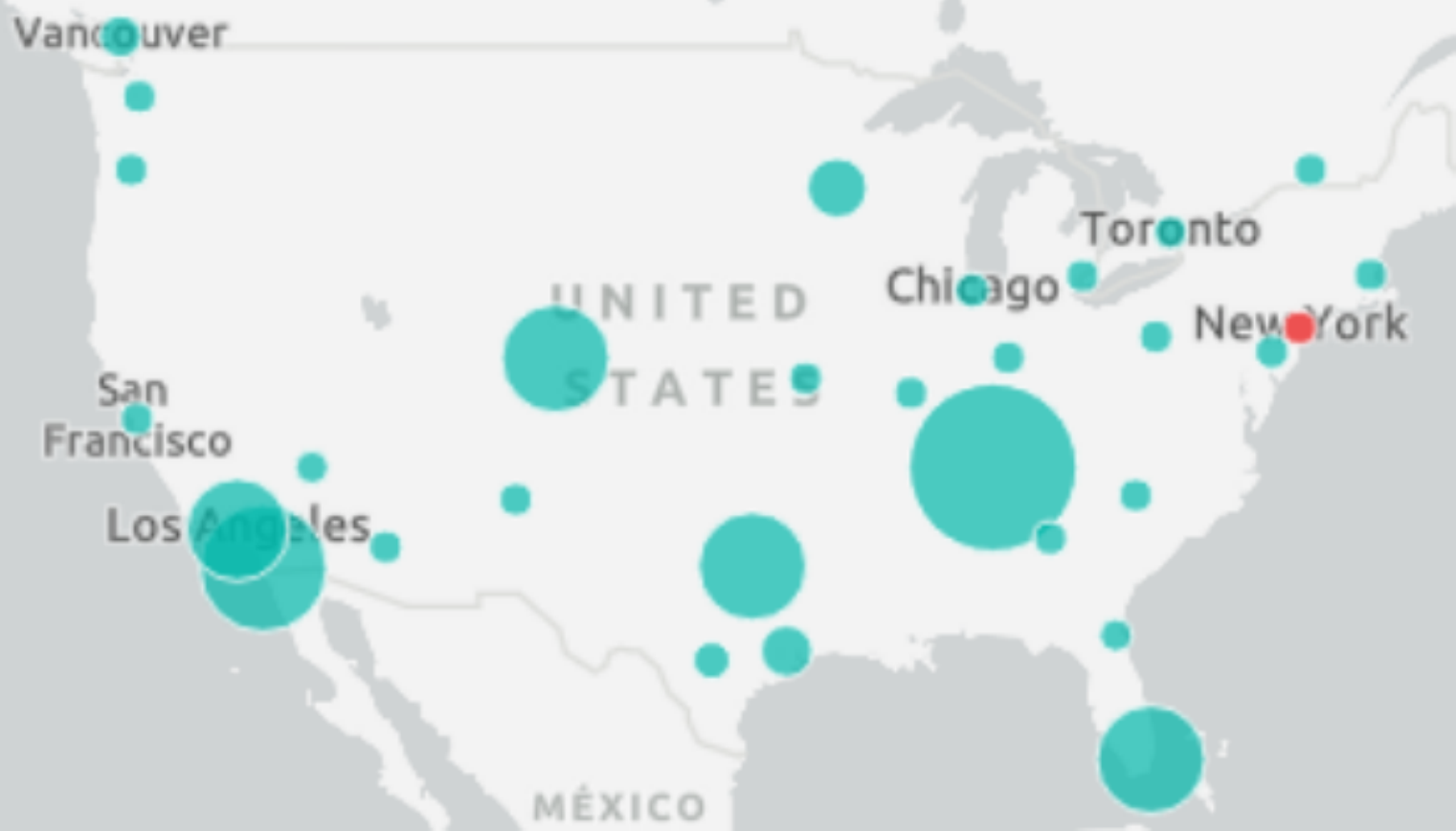}
    \caption{}
    \label{fig:NewYorkMaps_lag4}
  \end{subfigure}\hspace{0.001\textwidth}
  \begin{subfigure}{0.24\linewidth}
    \figuretitle{8 hours into the future}
    \includegraphics[width=\linewidth]{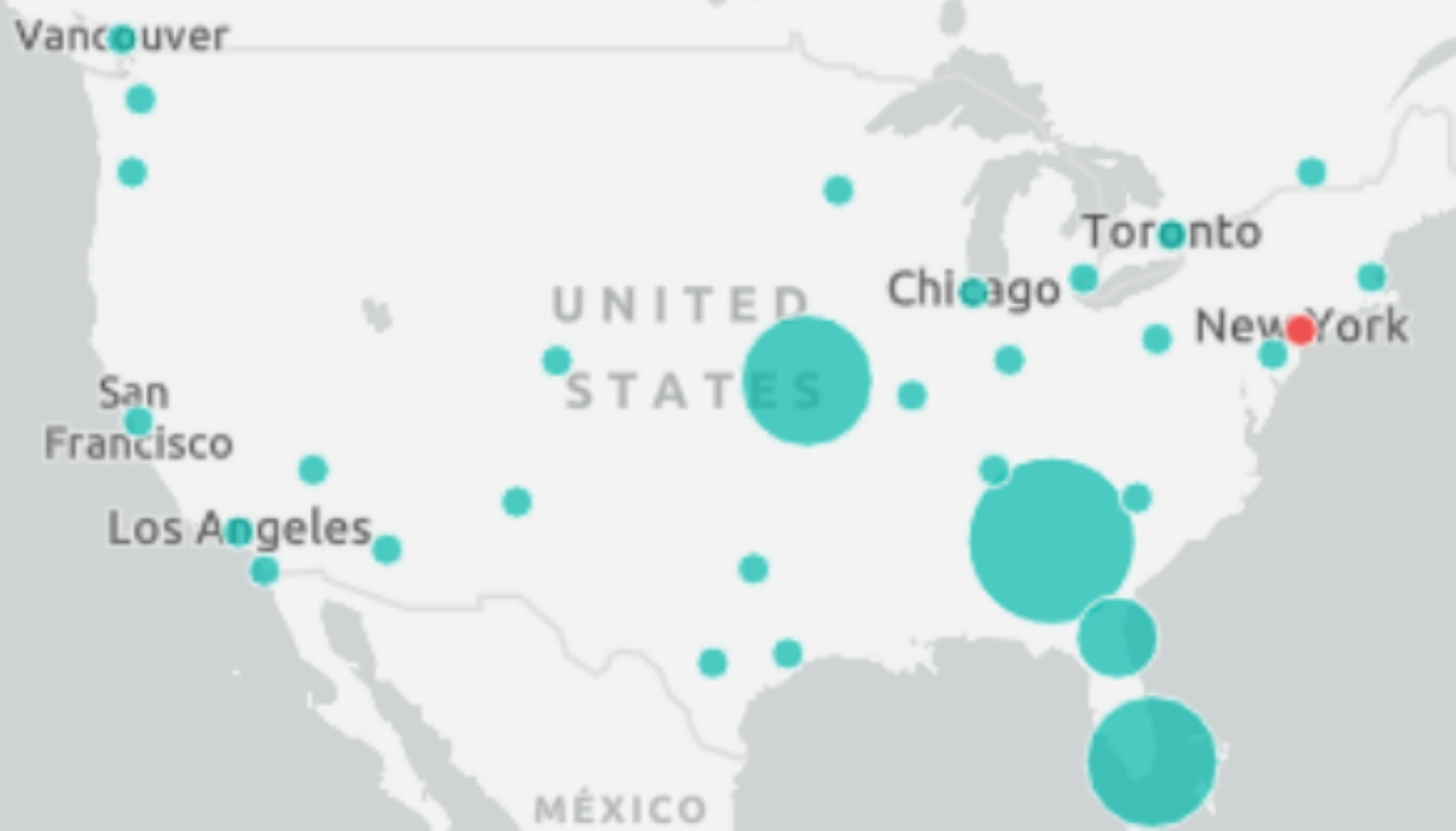}
    \caption{}
    \label{fig:NewYorkMaps_lag8}
  \end{subfigure}\hspace{0.001\textwidth}
  \begin{subfigure}{0.24\linewidth}
    \figuretitle{12 hours into the future}
    \includegraphics[width=\linewidth]{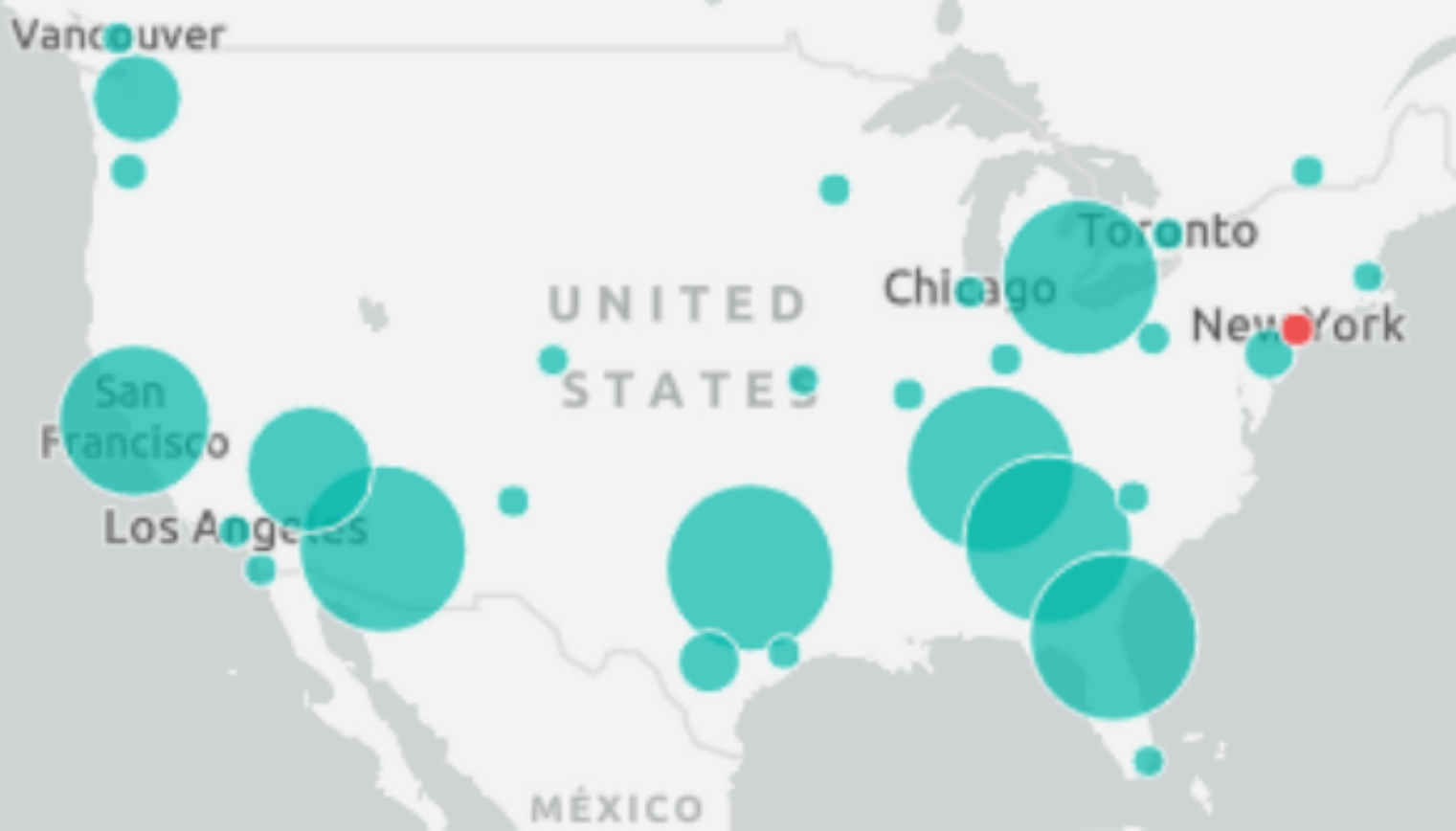}
    \caption{}
    \label{fig:NewYorkMaps_lag12}
  \end{subfigure}\hspace{0.001\textwidth}
  \begin{subfigure}{0.24\linewidth}
    \figuretitle{16 hours into the future}
    \includegraphics[width=\linewidth]{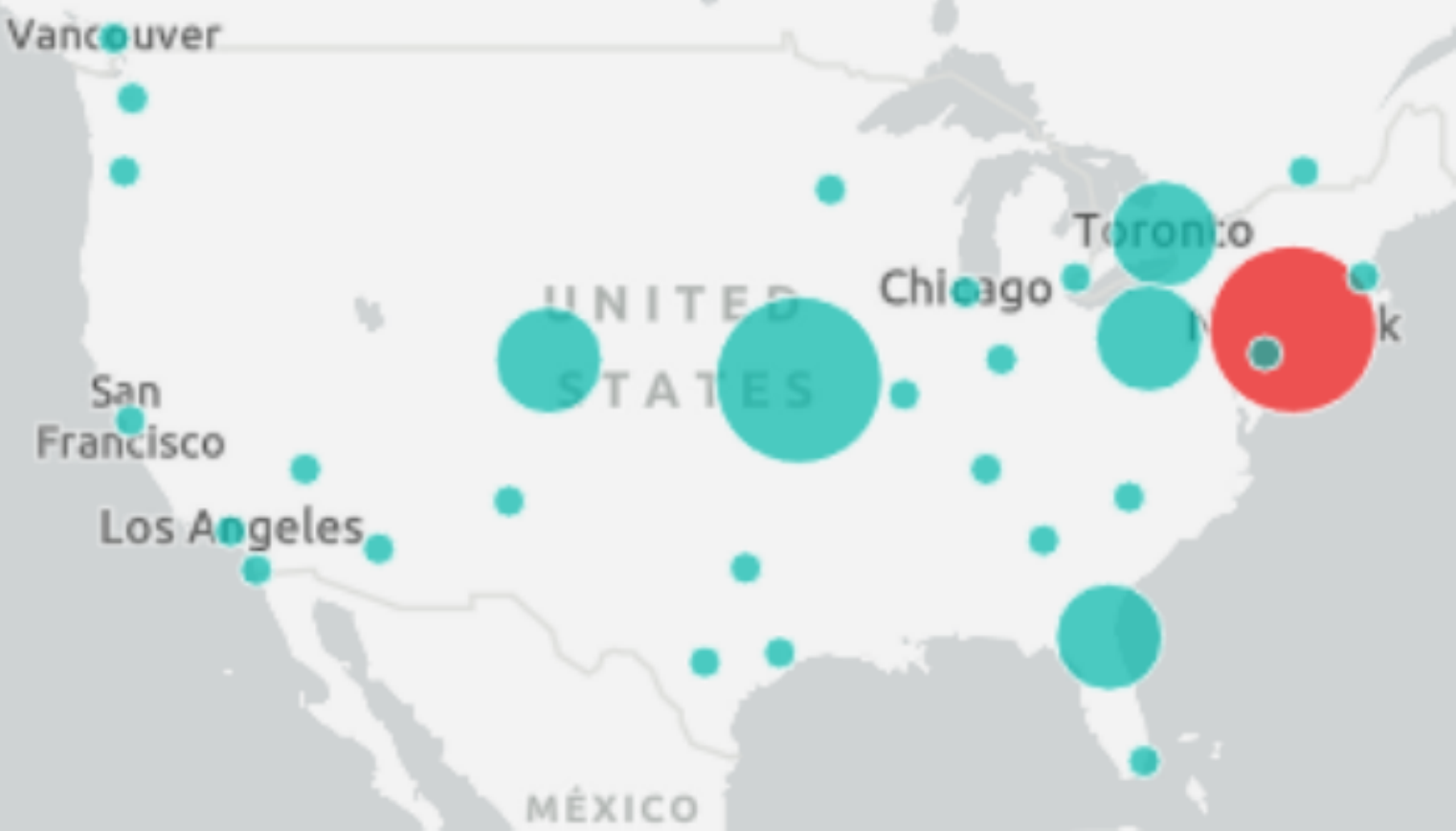}
    \caption{}
    \label{fig:NewYorkMaps_lag16}
  \end{subfigure}
  \caption{Attention visualization for \textbf{New York} in USA-Canada dataset. The circular graphs shows which city each of the most important heads attends to. The thickness of the line represents the amount of attention each of the heads is paying to the cities. The size of the circles indicates the importance of
each city in the temperature prediction for the target city. The target city is marked as a red circle and its size corresponds to the importance of the attention to itself.}
  \label{fig:NewYork}
\end{figure*}

\subsection{Attention Visualization}
The attention scores $AS_{c}^{h}$ in Eq. (\ref{eq:s_hc}) are used to show to which city each of the heads is paying attention to when predicting the output of the model. In particular, the analysis is provided for three target cities, i.e. Dallas, Vancouver and New York in Figs. \ref{fig:Dallas}, \ref{fig:Vancouver} and \ref{fig:NewYork} (a)-(d), respectively. In addition, in order to quantify the contribution of each city to the target city we compute $AS_{c}$ from Eq. (\ref{eq:s_c}). The visualization of the attention for the above-mentioned target cities are shown in Figs. \ref{fig:Dallas},  \ref{fig:Vancouver} and \ref{fig:NewYork} (e)-(h).  

In these plots, e.g. Fig. \ref{fig:Dallas} (a)-(d), one can see the name of cities and numbers, the latter represents each of the heads used in the Tensorial multi-head attention. The thickness of the connecting lines represents the amount of attention that each head gives to the city it is connected to. For the purpose of readability of the plots, we only keep the cities that receive the most attention and the heads that contributed to those cities. In Fig \ref{fig:Dallas} (e)-(h), the size of the circles indicates the importance of each city in the temperature prediction for the target city. The target city is shown as a red circle and its size indicates the importance of the attention to itself. It can be seen that in general the farther away we are predicting into the future, the distance between the target city and the most important cities for the prediction also increases. In general, a similar pattern as those of Fig. \ref{fig:Dallas} can be observed for Fig. \ref{fig:Vancouver} and \ref{fig:NewYork}.

\section{Conclusion}\label{Conclusion}

In this paper, we introduced TENT, a novel transformer-based model equipped with tensorial attention, for the temperature prediction task. The model is tested on two real-life weather datasets and the it outperforms the other examined models in USA-Canada dataset . In the EU dataset, our model is in general the second best architecture. Two attention scores are introduced to and visualized in order to provide additional insights on the model predictions.   
The introduced model can potentially be used for other applications with 3D tensor inputs and our code\footnote{\url{https://github.com/onurbil/TENT}} is available online.

\bibliography{mybibfile}

\begin{thebibliography}{10}
\expandafter\ifx\csname url\endcsname\relax
  \def\url#1{\texttt{#1}}\fi
\expandafter\ifx\csname urlprefix\endcsname\relax\def\urlprefix{URL }\fi
\expandafter\ifx\csname href\endcsname\relax
  \def\href#1#2{#2} \def\path#1{#1}\fi

\bibitem{lai2018modeling}
G.~Lai, W.-C. Chang, Y.~Yang, H.~Liu, Modeling long-and short-term temporal
  patterns with deep neural networks, in: The 41st International ACM SIGIR
  Conference on Research \& Development in Information Retrieval, 2018, pp.
  95--104.

\bibitem{MEHRKANOON_RepresLearning}
S.~Mehrkanoon, Deep shared representation learning for weather elements
  forecasting, Knowledge-Based Systems 179 (2019) 120--128.

\bibitem{trebing2020wind}
K.~Trebing, S.~Mehrkanoon, Wind speed prediction using multidimensional
  convolutional neural networks, in: IEEE Symposium Series on Computational
  Intelligence (IEEE-SSCI), IEEE, 2020, pp. 713--720.

\bibitem{fernandez2020deep}
J.~G. Fern{\'a}ndez, I.~A. Abdellaoui, S.~Mehrkanoon, Deep coastal sea elements
  forecasting using \uppercase{u}-\uppercase{n}et based models, arXiv preprint
  arXiv:2011.03303 (2020).

\bibitem{trebing2021smaat}
K.~Trebing, T.~Sta{\'n}czyk, S.~Mehrkanoon, Sma\uppercase{a}t-\uppercase{un}et:
  Precipitation nowcasting using a small attention-unet architecture, Pattern
  Recognition Letters 145 (2021) 178--186.

\bibitem{gcn_t}
T.~Stanczyk, S.~Mehrkanoon, Deep graph convolutional networks for wind speed
  prediction, in: European Symposium on Artificial Neural Networks,
  Computational Intelligence and Machine Learning (ESANN), 2021, pp. 147--152.

\bibitem{aykas2021multistream}
D.~Aykas, S.~Mehrkanoon, Multistream graph attention networks for wind speed
  forecasting, in: IEEE Symposium Series on Computational Intelligence
  (IEEE-SSCI), IEEE, 2021, pp. 1--8.

\bibitem{Ismail21}
I.~A. Abdellaoui, S.~Mehrkanoon, Symbolic regression for scientific discovery:
  an application to wind speed forecasting, in: 2021 IEEE Symposium Series on
  Computational Intelligence (SSCI), 2021, pp. 01--08.
\newblock \href {https://doi.org/10.1109/SSCI50451.2021.9659860}
  {\path{doi:10.1109/SSCI50451.2021.9659860}}.

\bibitem{fernandez2021broad}
J.~G. Fern{\'a}ndez, S.~Mehrkanoon, Broad-\uppercase{UN}et: Multi-scale feature
  learning for nowcasting tasks, Neural Networks 144 (2021) 419--427.

\bibitem{kreuzer2020short}
D.~Kreuzer, M.~Munz, S.~Schl{\"u}ter, Short-term temperature forecasts using a
  convolutional neural network—an application to different weather stations
  in germany, Machine Learning with Applications 2 (2020) 100007.

\bibitem{ahmad2020maximizing}
S.~K. Ahmad, F.~Hossain, Maximizing energy production from hydropower dams
  using short-term weather forecasts, Renewable Energy 146 (2020) 1560--1577.

\bibitem{dai2006tensor}
G.~Dai, D.-Y. Yeung, Tensor embedding methods, in: AAAI, Vol.~6, 2006, pp.
  330--335.

\bibitem{nguyen2015tensor}
T.~D. Nguyen, T.~Tran, D.~Phung, S.~Venkatesh, Tensor-variate restricted
  boltzmann machines, in: Twenty-Ninth AAAI Conference on Artificial
  Intelligence, 2015.

\bibitem{ma2019tensorized}
X.~Ma, P.~Zhang, S.~Zhang, N.~Duan, Y.~Hou, D.~Song, M.~Zhou, A tensorized
  transformer for language modeling (2019).
\newblock \href {http://arxiv.org/abs/1906.09777} {\path{arXiv:1906.09777}}.

\bibitem{marchuk2012numerical}
G.~Marchuk, Numerical methods in weather prediction, Elsevier, 2012.

\bibitem{richardson2007weather}
L.~F. Richardson, Weather prediction by numerical process, Cambridge university
  press, 2007.

\bibitem{soman2010review}
S.~S. Soman, H.~Zareipour, O.~Malik, P.~Mandal, A review of wind power and wind
  speed forecasting methods with different time horizons, in: North American
  Power Symposium 2010, IEEE, 2010, pp. 1--8.

\bibitem{bauer2015quiet}
P.~Bauer, A.~Thorpe, G.~Brunet, The quiet revolution of numerical weather
  prediction, Nature 525~(7567) (2015) 47--55.

\bibitem{ravuri2021skilful}
S.~Ravuri, K.~Lenc, M.~Willson, D.~Kangin, R.~Lam, P.~Mirowski, M.~Fitzsimons,
  M.~Athanassiadou, S.~Kashem, S.~Madge, et~al., Skilful precipitation
  nowcasting using deep generative models of radar, Nature 597~(7878) (2021)
  672--677.

\bibitem{chen2011comparison}
L.~Chen, X.~Lai, Comparison between arima and ann models used in short-term
  wind speed forecasting, in: 2011 Asia-Pacific Power and Energy Engineering
  Conference, IEEE, 2011, pp. 1--4.

\bibitem{kuligowski1998localized}
R.~J. Kuligowski, A.~P. Barros, Localized precipitation forecasts from a
  numerical weather prediction model using artificial neural networks, Weather
  and forecasting 13~(4) (1998) 1194--1204.

\bibitem{bartos2006nonlinear}
I.~Bartos, I.~M. J\'anosi, Nonlinear correlations of daily temperature records
  over land, Nonlinear Processes in Geophysics 13~(5) (2006) 571--576.

\bibitem{salman2015weather}
A.~G. Salman, B.~Kanigoro, Y.~Heryadi, Weather forecasting using deep learning
  techniques, in: 2015 International Conference on Advanced Computer Science
  and Information Systems (ICACSIS), IEEE, 2015, pp. 281--285.

\bibitem{cifuentes2020air}
J.~Cifuentes, G.~Marulanda, A.~Bello, J.~Reneses, Air temperature forecasting
  using machine learning techniques: a review, Energies 13~(16) (2020) 4215.

\bibitem{klein2015dynamic}
B.~Klein, L.~Wolf, Y.~Afek, A dynamic convolutional layer for short range
  weather prediction, in: Proceedings of the IEEE Conference on Computer Vision
  and Pattern Recognition, 2015, pp. 4840--4848.

\bibitem{hochreiter1997long}
S.~Hochreiter, J.~Schmidhuber, Long short-term memory, Neural computation 9~(8)
  (1997) 1735--1780.

\bibitem{shi2015convolutional}
X.~Shi, Z.~Chen, H.~Wang, D.-Y. Yeung, W.-K. Wong, W.-c. Woo, Convolutional
  lstm network: A machine learning approach for precipitation nowcasting, arXiv
  preprint arXiv:1506.04214 (2015).

\bibitem{vaswani2017attention}
A.~Vaswani, N.~Shazeer, N.~Parmar, J.~Uszkoreit, L.~Jones, A.~N. Gomez,
  {\L}.~Kaiser, I.~Polosukhin, Attention is all you need, in: Advances in
  neural information processing systems, 2017, pp. 5998--6008.

\bibitem{shih2019temporal}
S.-Y. Shih, F.-K. Sun, H.-y. Lee, Temporal pattern attention for multivariate
  time series forecasting, Machine Learning 108~(8-9) (2019) 1421--1441.

\bibitem{qin2017dual}
Y.~Qin, D.~Song, H.~Chen, W.~Cheng, G.~Jiang, G.~Cottrell, A dual-stage
  attention-based recurrent neural network for time series prediction, arXiv
  preprint arXiv:1704.02971 (2017).

\bibitem{bahdanau2014neural}
D.~Bahdanau, K.~Cho, Y.~Bengio, Neural machine translation by jointly learning
  to align and translate, arXiv preprint arXiv:1409.0473 (2014).

\bibitem{liu2018generating}
P.~J. Liu, M.~Saleh, E.~Pot, B.~Goodrich, R.~Sepassi, L.~Kaiser, N.~Shazeer,
  Generating wikipedia by summarizing long sequences, arXiv preprint
  arXiv:1801.10198 (2018).

\bibitem{kitaev2018constituency}
N.~Kitaev, D.~Klein, Constituency parsing with a self-attentive encoder, arXiv
  preprint arXiv:1805.01052 (2018).

\bibitem{devlin2018bert}
J.~Devlin, M.-W. Chang, K.~Lee, K.~Toutanova, Bert: Pre-training of deep
  bidirectional transformers for language understanding, arXiv preprint
  arXiv:1810.04805 (2018).

\bibitem{radford2018improving}
A.~Radford, K.~Narasimhan, T.~Salimans, I.~Sutskever, Improving language
  understanding with unsupervised learning, Technical report, OpenAI (2018).

\bibitem{parmar2018image}
N.~Parmar, A.~Vaswani, J.~Uszkoreit, {\L}.~Kaiser, N.~Shazeer, A.~Ku, D.~Tran,
  Image transformer, arXiv preprint arXiv:1802.05751 (2018).

\bibitem{huang2018music}
C.-Z.~A. Huang, A.~Vaswani, J.~Uszkoreit, N.~Shazeer, I.~Simon, C.~Hawthorne,
  A.~M. Dai, M.~D. Hoffman, M.~Dinculescu, D.~Eck, Music transformer, arXiv
  preprint arXiv:1809.04281 (2018).

\bibitem{shen2018tensorized}
T.~Shen, T.~Zhou, G.~Long, J.~Jiang, C.~Zhang, Tensorized self-attention:
  Efficiently modeling pairwise and global dependencies together, arXiv
  preprint arXiv:1805.00912 (2018).

\bibitem{shen2018disan}
T.~Shen, T.~Zhou, G.~Long, J.~Jiang, S.~Pan, C.~Zhang, Disan: Directional
  self-attention network for rnn/cnn-free language understanding, in:
  Proceedings of the AAAI conference on artificial intelligence, Vol.~32, 2018.

\bibitem{zhou2021informer}
H.~Zhou, S.~Zhang, J.~Peng, S.~Zhang, J.~Li, H.~Xiong, W.~Zhang, Informer:
  Beyond efficient transformer for long sequence time-series forecasting, in:
  Proceedings of AAAI, 2021.

\bibitem{tucker1966some}
L.~R. Tucker, Some mathematical notes on three-mode factor analysis,
  Psychometrika 31~(3) (1966) 279--311.

\bibitem{li2017bt}
G.~Li, J.~Ye, H.~Yang, D.~Chen, S.~Yan, Z.~Xu, Bt-nets: Simplifying deep neural
  networks via block term decomposition, arXiv preprint arXiv:1712.05689
  (2017).

\bibitem{de2008decompositions}
L.~De~Lathauwer, Decompositions of a higher-order tensor in block terms—part
  ii: Definitions and uniqueness, SIAM Journal on Matrix Analysis and
  Applications 30~(3) (2008) 1033--1066.

\bibitem{wiegreffe2019attention}
S.~Wiegreffe, Y.~Pinter, Attention is not not explanation, arXiv preprint
  arXiv:1908.04626 (2019).

\bibitem{Gui_Ge_Hu_2019}
N.~Gui, D.~Ge, Z.~Hu,
  \href{https://ojs.aaai.org/index.php/AAAI/article/view/4255}{Afs: An
  attention-based mechanism for supervised feature selection}, Proceedings of
  the AAAI Conference on Artificial Intelligence 33~(01) (2019) 3705--3713.
\newblock \href {https://doi.org/10.1609/aaai.v33i01.33013705}
  {\path{doi:10.1609/aaai.v33i01.33013705}}.
\newline\urlprefix\url{https://ojs.aaai.org/index.php/AAAI/article/view/4255}

\bibitem{LSTM}
S.~Hochreiter, J.~Schmidhuber, Long short-term memory, Neural Computation 9
  (1997) 1735--1780.
\newblock \href {https://doi.org/10.1162/neco.1997.9.8.1735}
  {\path{doi:10.1162/neco.1997.9.8.1735}}.

\bibitem{ConvLSTM}
X.~SHI, Z.~Chen, H.~Wang, D.-Y. Yeung, W.-k. Wong, W.-c. WOO,
  \href{https://proceedings.neurips.cc/paper/2015/file/07563a3fe3bbe7e3ba84431ad9d055af-Paper.pdf}{Convolutional
  lstm network: A machine learning approach for precipitation nowcasting}, in:
  C.~Cortes, N.~Lawrence, D.~Lee, M.~Sugiyama, R.~Garnett (Eds.), Advances in
  Neural Information Processing Systems, Vol.~28, Curran Associates, Inc.,
  2015.
\newline\urlprefix\url{https://proceedings.neurips.cc/paper/2015/file/07563a3fe3bbe7e3ba84431ad9d055af-Paper.pdf}

\bibitem{kingma2014adam}
D.~P. Kingma, J.~Ba, Adam: A method for stochastic optimization, arXiv preprint
  arXiv:1412.6980 (2014).

\bibitem{claessens2019efficient}
S.~J. Claessens, Efficient transformation from cartesian to geodetic
  coordinates, Computers \& Geosciences 133 (2019) 104307.

\bibitem{hofmann2006physical}
B.~Hofmann-Wellenhof, H.~Moritz, Physical geodesy, Springer Science \& Business
  Media, 2006.

\bibitem{guo2019star}
Q.~Guo, X.~Qiu, P.~Liu, Y.~Shao, X.~Xue, Z.~Zhang, Star-transformer, arXiv
  preprint arXiv:1902.09113 (2019).

\bibitem{ezen2020comparison}
A.~Ezen-Can, A comparison of lstm and bert for small corpus, arXiv preprint
  arXiv:2009.05451 (2020).

\end{thebibliography}
\end{document}